\documentclass[acmsmall]{acmart}
\AtBeginDocument{%
  }

\setcopyright{acmlicensed}
\copyrightyear{2025}
\acmYear{2025}
\acmDOI{XXXXXXX.XXXXXXX}

\usepackage{longtable}
\usepackage{caption}    
\usepackage{booktabs,makecell, multirow, tabularx}
\usepackage{array} 
\usepackage{tikz}
\usetikzlibrary{positioning, shapes, arrows.meta}
\usepackage{xcolor}

\definecolor{mainblue}{RGB}{173, 235, 242}

\definecolor{red1}{RGB}{255, 235, 235}
\definecolor{red2}{RGB}{255, 178, 178}
\definecolor{red3}{RGB}{255, 128, 128}

\definecolor{orange1}{RGB}{255, 245, 225}
\definecolor{orange2}{RGB}{255, 204, 153}
\definecolor{orange3}{RGB}{255, 170, 102}

\definecolor{green1}{RGB}{204, 255, 204}
\definecolor{green2}{RGB}{170, 240, 170}
\definecolor{green3}{RGB}{140, 220, 140}

\definecolor{textgray}{RGB}{240,240,240}

\begin{document}


\title{Reinforced MLLM: A Survey on RL-Based Reasoning in Multimodal Large Language Models}

\author{Guanghao Zhou}
\authornote{Both authors contributed equally to this research.}
\email{ghzhou@stu.ecnu.edu.cn}
\author{Panjia Qiu}
\authornotemark[1]
\email{panjiaqiu@stu.ecnu.edu.cn}
\affiliation{%
  \institution{East China Normal University}
  \city{Shanghai}
  \country{China}
}

\author{Cen Chen}
\authornote{Corresponding author.}
\affiliation{%
  \institution{East China Normal University}
  \city{Shanghai}
  \country{China}}
\email{cenchen@dase.ecnu.edu.cn}

\author{Jie Wang}
\affiliation{
  \institution{ByteDance}
  \city{Shanghai}
  \country{China}
}
\email{wangjie.mayday@bytedance.com}

\author{Zheming Yang}
\affiliation{%
  \institution{ByteDance}
  \city{Beijing}
  \country{China}
}
\email{yangzheming@bytedance.com}

\author{Jian Xu}
\affiliation{%
 \institution{ByteDance}
 \city{Shanghai}
 \country{China}}
\email{kid44106592@gmail.com}

\author{Minghui Qiu}
\affiliation{%
 \institution{ByteDance}
 \city{Shanghai}
 \country{China}}
\email{minghuiqiu@gmail.com}

\renewcommand{\shortauthors}{Guanghao Zhou, Panjia Qiu et al.}
\newcommand{\qiu}[1]{{\color{blue}[\textbf{\sc qiu}: #1]}}
\newcommand{\zhou}[1]{{\color{orange}[\textbf{\sc zhou}: #1]}}
\newcommand{\jie}[1]{{\color{green}[\textbf{\sc jie}: #1]}}
\newcommand{\panjia}[1]{{\color{purple}[\textbf{\sc panjia}: #1]}}

\begin{abstract}
The application of reinforcement learning (RL) to enhance the reasoning capabilities of Multimodal Large Language Models (MLLMs) constitutes a rapidly advancing research area. While MLLMs extend Large Language Models (LLMs) to handle diverse modalities such as vision, audio, and video, enabling robust reasoning across multimodal inputs remains challenging. This paper provides a systematic review of recent advances in RL-based reasoning for MLLMs, covering key algorithmic designs, reward mechanism innovations, and practical applications. We highlight two main RL paradigms, value-model-free and value-model-based methods, and analyze how RL enhances reasoning abilities by optimizing reasoning trajectories and aligning multimodal information. Additionally, we provide an extensive overview of benchmark datasets, evaluation protocols, and current limitations, and propose future research directions to address challenges such as sparse rewards, inefficient cross-modal reasoning, and real-world deployment constraints. Our goal is to provide a comprehensive and structured guide to RL-based multimodal reasoning.
\end{abstract}



\received{20 February 2007}
\received[revised]{12 March 2009}
\received[accepted]{5 June 2009}

\maketitle

\section{Introduction}

The emergence of large language models (LLMs) \cite{yang2024qwen2, glm2024chatglm, grattafiori2024llama} has ushered in an unprecedented new era for the field of artificial intelligence, showcasing exceptional capabilities such as instruction following and few-shot learning \cite{brown2020language}. However, achieving human-level intelligence requires not only surpassing basic perceptual abilities but also developing sophisticated cognitive reasoning skills that enable iterative reasoning through contextual understanding and self-correction. Inspired by this, in-context learning (ICL) techniques \cite{wei2022emergent, wu2022self} have empowered LLMs with step-by-step reasoning abilities, commonly referred to as the Chain-of-Thought (CoT) reasoning mechanism \cite{wang2024chain, besta2024graph}. 
OpenAI's o1 model \cite{jaech2024openai} has demonstrated remarkable performance in solving reasoning tasks, sparking widespread interest across domains in test-time scaling research. By leveraging additional computation during inference time to conduct "slow thinking" \cite{kirillov2023segment}, it further enhances the accuracy of responses to complex problems.

\begin{figure}[t!]
    \centering
    \includegraphics[width=0.85\columnwidth]{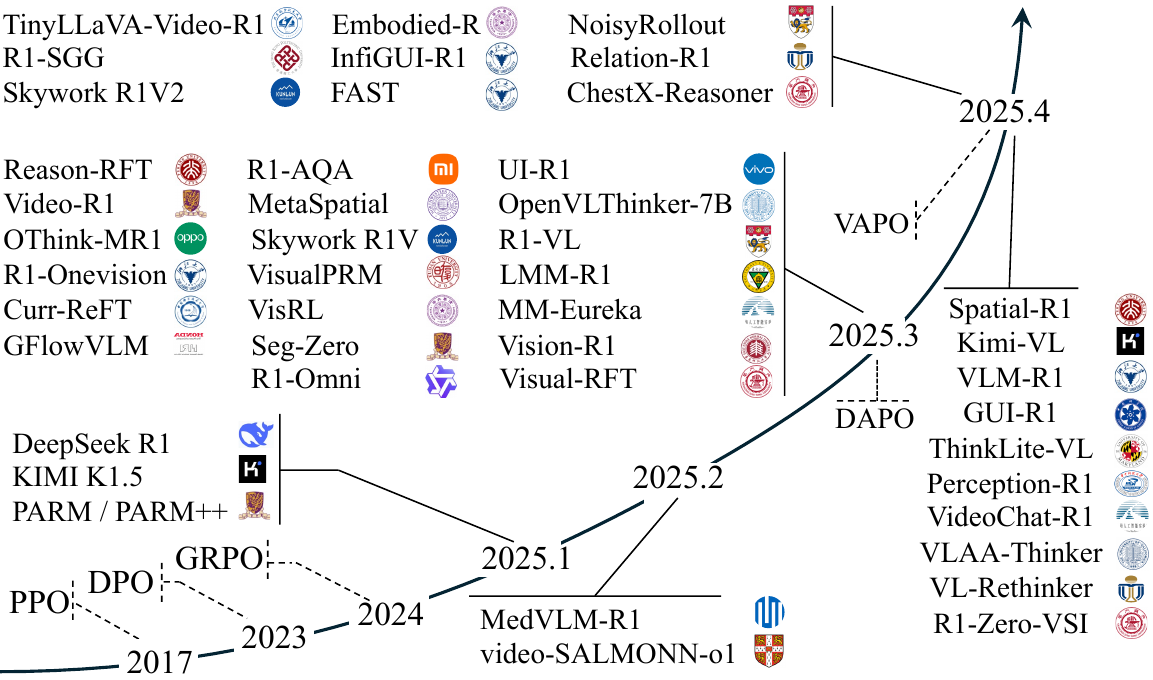}
    \caption{Timeline of RL-based reasoning methods for MLLM.}
    \label{fig:timeline}
\end{figure}

Inspired by the extensive research on the CoT in LLMs, reasoning tasks in multi-modal large language models (MLLMs) \cite{wang2024qwen2, bai2025qwen2, team2025kimi} have experienced rapid advancements. Typical methods include Best-of-N, Beam Search, and Monte Carlo Tree Search \cite{xu2024llava, yao2024mulberry, chen2024alphamath}. These approaches rely on intricate search mechanisms to generate large volumes of reasoning data and utilize supervised fine-tuning (SFT) to enable models to acquire autonomous reasoning capabilities.
With the advancement of reinforcement learning (RL) theory and technology, DeepSeek R1 \cite{guo2025deepseek} demonstrates how large language models can autonomously develop complex reasoning abilities through simple rule-based incentive mechanisms and lightweight reinforcement learning (GRPO \cite{shao2024deepseekmath}) algorithms.
This method enables LLMs to naturally exhibit "Aha Moment" without explicit supervision, characterized by self-reflection and spontaneous increases in response length during training. Recent studies \cite{huang2025vision, zhou2025r1, liu2025visual, pan2025medvlm} have extended this approach to MLLMs and applied it to areas such as object recognition \cite{liu2025visual}, semantic segmentation \cite{liu2025seg}, and video analysis \cite{sun2025video}. These methods significantly enhance MLLMs' capabilities with limited training data, achieving performance comparable to SFT in in-domain tests while outperforming SFT models in out-of-distribution (OOD) evaluations. However, this rapid progress, as illustrated in Figure \ref{fig:timeline}, presents significant challenges to researchers. Although RL-based methods are effective, most of them adhere to text-based thinking practices, neglecting the critical role of other modalities in multimodal scenarios. Moreover, current RL-based reasoning methods primarily rely on rule-based reward functions with verifiable answers, overlooking broader general-scenario problems where verifiable answers are unavailable.

Although many existing surveys focus on MLLMs reasoning \cite{wang2025multimodal, lin2025investigating}, none of them specifically address the issue of RL-based reasoning for MLLMs. 
To fill this gap, we provide a comprehensive and systematic review of reasoning methods for RL-based MLLMs, offering a structured analysis of technological advancements, methodologies, practical applications, and future directions. Our goal is to provide researchers with a thorough and systematic guide to identifying appropriate methods in the rapidly evolving field of MLLM reasoning, thereby promoting further innovation and progress in this dynamic area. We first introduce the background of MLLMs, CoT reasoning, and reinforcement learning in Section \ref{sec:backAndPrelim}. Then, we review key RL algorithm designs and optimizations for both LLMs and MLLMs in Section \ref{sec:RL}. Next, we survey RL-based reasoning methods for MLLMs, covering algorithmic strategies, reward mechanisms, and benchmark evaluations in Section \ref{sec:MLLMReasoning}, Section \ref{sec:application} and Section \ref{sec:dataAndBench}. Finally, Section \ref{sec:limAndFuture} discuss current limitations and future research directions in this rapidly evolving field in. 

This paper addresses RL-based reasoning for MLLMs from four key perspectives:

\begin{itemize}
    \item \textbf{Core Designs of RL in LLMs/MLLMs} : Analyze value-model-free and value-model-based methods, focusing on their innovations in training efficiency, stability, and performance, while discussing strengths, weaknesses, and optimization opportunities.
    \item \textbf{Analysis of RL-Based Reasoning Methods} : Examine the algorithmic frameworks, the design of reward functions including both accuracy-oriented and structure-oriented approaches, and the strategies for integrating multiple modalities.  Furthermore, emphasize how each of these components contributes to addressing the core challenges of multimodal reasoning.
    \item \textbf{Benchmark Datasets and Evaluation Protocols} : Survey datasets and evaluation methods for reasoning tasks, covering construction pipelines such as data sources and annotations, as well as benchmarks spanning math, science, spatial, and interaction-based domains.
    \item \textbf{Limitations and Future Directions} : Highlight unresolved challenges such as sparse rewards, inefficient trajectories, and cross-modal coordination. Meanwhile, propose future directions including hierarchical reward modeling, vision-guided CoT, and lightweight RL frameworks for real-world use.
\end{itemize}

\section{Background and Preliminary}
\label{sec:backAndPrelim}
\subsection{Multi-Modal Large Language Models and MM-COT}

Despite the impressive zero-shot/few-shot reasoning performance of LLMs in most Natural Language Processing (NLP) tasks, they are limited to understanding only discrete textual information. In contrast, models for other modalities are capable of integrating information from various modalities such as images, audio, and video \cite{bai2025qwen2, li2025reinforcement, sun2025video}, but often fall short in reasoning capabilities. Given this complementarity, the integration of LLMs with other specialized models has led to the emergence of multimodal large language models. 
Representative examples include open-source models such as Qwen-VL \cite{wang2024qwen2, bai2025qwen2}, Intern-VL \cite{chen2024internvl, zhu2025internvl3}, LLaVA \cite{liu2023visual, liu2024improved}, and Valley \cite{luo2023valley, wu2025valley2}, as well as closed-models like GPT-4o \cite{hurst2024gpt} and Gemini \cite{team2023gemini,  team2024gemini}. 
These MLLMs leverage LLMs as the core cognitive engine, demonstrating exceptional capabilities such as strong language generation, zero-shot transfer, and ICL. At the same time, foundational models for other modalities empower a variety of multimodal tasks by providing high-quality representations for non-textual data.
In summary, mainstream MLLMs typically follow a consistent architecture: they process multimodal embeddings or tokens through a decoder structure, generating context-aware outputs autoregressively. Some studies have attempted to enable MLLMs to handle arbitrary modality conversions \cite{bai2025qwen2, xu2025qwen2}, further advancing the development of "anything-to-anything" paradigm models.

\begin{figure}[t!]
    \centering
    \includegraphics[width=\columnwidth]{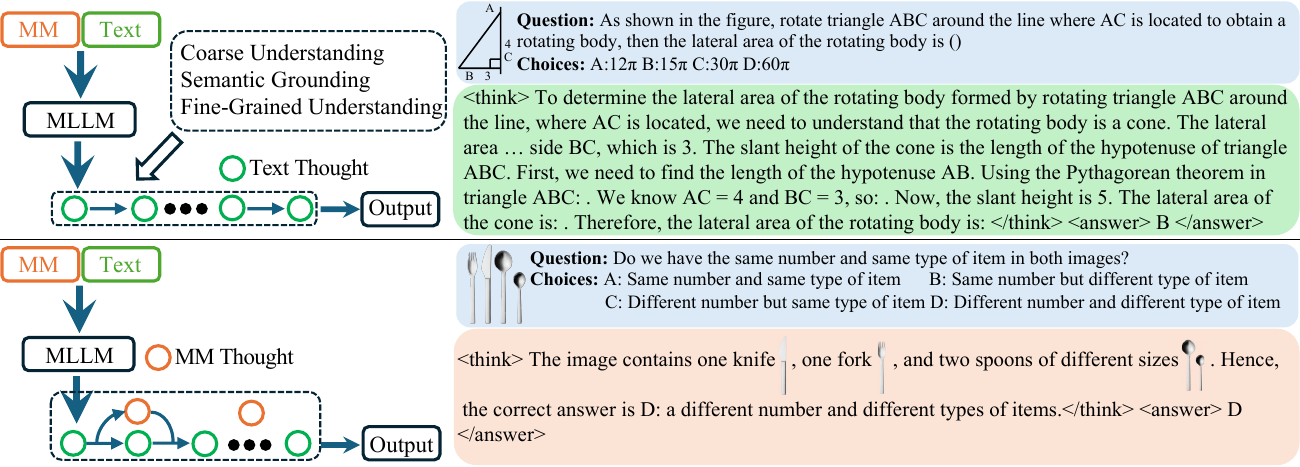}
    \caption{Illustration of Different Modalities of CoT. The upper subplot shows a text-only CoT, while the lower subplot presents a multimodal CoT.}
    \label{fig:mm-cot}
\end{figure}

CoT is a technique that encourages models to generate intermediate reasoning steps $S = (s_1, \dots, s_n)$ before arriving at the final answer $A$. In the context of multimodal reasoning tasks, given a prompt $P$ and query $Q$, each reasoning step $s_i$ is sampled from the MLLM $\mathcal{M}$:
\begin{align}
    p(S|P, Q)=   \prod_{i=1}^{|S|} \prod_{j=1}^{|s_i|}  \mathcal{M}(s_i|P, Q, S_{<i}, s_{i<j}) \\
    P(A|P, Q, S) = \prod_{i=1}^{|A|} \mathcal{M}(a_i | P, Q, S, a_{<i})
\end{align}
Unlike traditional CoT, Multimodal Chain-of-Thought (MM-CoT) attempts to incorporate multimodal information into $P$, $S$, and $A$ to varying degrees. As illustrated in Figure \ref{fig:mm-cot}, we categorize MM-CoT into two distinct scenarios: one that relies solely on linguistic information and another that integrates multimodal signals beyond textual content. The former aims to solve tasks involving multi-modal inputs using purely text-based CoT, while the latter emphasizes the integration of given, retrieved, or generated multi-modal information within the MM-CoT itself. Since most popular MLLMs still struggle to generate images or other modalities, recent advancements in RL-based reasoning have primarily concentrated on the generation of text-only CoT.

\subsection{Reinforcement Learning}

\subsubsection{Policy Optimization Methods}

Reinforcement Learning is a fundamental approach in machine learning where an agent learns to interact with an environment by taking actions, receiving rewards, and updating its policy to maximize long-term returns. Reinforcement Learning from Human Feedback (RLHF) \cite{bai2022training}, leveraging algorithms such as Proximal Policy Optimization (PPO) \cite{schulman2017proximal}, guides model behavior to enhance the alignment, coherence, and utility of generated responses. 
\paragraph{Proximal Policy Optimization (PPO)} PPO optimizes LLMs by maximizing the following surrogate objective:
\begin{align}
J_{PPO}(\theta)&=\mathbb{E}[q\sim P(Q),o\sim\pi_{\theta_{old}}(O|q)] \\ \notag
&\frac{1}{|o|}\sum_{t=1}^{|o|}\min\left[\frac{\pi_\theta(o_t|q,o_{<t})}{\pi_{\theta_{old}}(o_t|q,o_{<t})}A_t,\mathrm{clip}\left(\frac{\pi_\theta(o_t|q,o_{<t})}{\pi_{\theta_{old}}(o_t|q,o_{<t})},1-\varepsilon,1+\varepsilon\right)A_t\right]
\end{align}
where $\pi_\theta$ and $\pi_{\theta_{old}}$ denote the current and old policy models, respectively, $q$ and $o$ represent questions sampled from the question dataset and outputs generated by the old policy $\pi_{\theta_{old}}$. The hyperparameter $\varepsilon$ plays a critical role in stabilizing training by introducing a clipping mechanism. $A_t$, computed using Generalized Advantage Estimation (GAE) \cite{schulman2015high}, represents the advantage function based on a reward model $R$ and a learnable critic model.

In recent years, RL has increasingly been applied to enhance the reasoning capabilities of LLMs \cite{guo2025deepseek, team2025kimi}. By treating reasoning token generation as a Markov Decision Process, models are trained to maximize the expected return of reasoning paths, guiding optimization toward more structured and coherent reasoning trajectories. In PPO, both the policy and critic models must be trained simultaneously, which imposes significant computational demands when model parameters or token counts are large. 
To reduce the resource consumption of PPO while stabilizing the training process, some studies have begun exploring ways to simplify the PPO training pipeline.

\paragraph{REINFORCE Leave-One-Out (RLOO)} 
RLOO \cite{ahmadian2024back} omits the use of a critic model and GAE , instead directly employing the Monte Carlo method to compute the baseline. Specifically, for each query $x$, the model generates $G$ responses $\{o_{1}, o_{2}, \dots, o_{G}\}$ and calculates a score for each response $\{r_1, r_2, \dots, r_G\}$. RLOO uses a leave-one-out baseline to reduce the variance in policy gradient estimation, with the advantage estimate computed as follows:

\begin{align}
    A(i) = r(i) - \frac{1}{G-1} \sum_{j \neq i} r(j), \quad i = 1, \dots, G.
\end{align}

By modeling the entire process as a bandit problem, this method treats the entire response as an action, discarding the token-level rewards used in PPO and instead relying solely on sequence-level rewards. Although this method introduces higher variance, the authors consider it acceptable given the strong performance of pre-trained models.

\paragraph{Group Relative Policy Optimization (GRPO)} GRPO \cite{shao2024deepseekmath} eliminates the need for the critic model by directly comparing generated response groups. Its optimization objective is defined as:
\begin{align}
\label{equ:GRPO}
\mathcal{J}_{GRPO}(\theta) & =\mathbb{E}[q\sim P(Q),\{o_i\}_{i=1}^G\sim\pi_{\theta_{old}}(O|q)] \\ \notag
 & \frac{1}{G}\sum_{i=1}^G\left(\min\left(\frac{\pi_\theta(o_i|q)}{\pi_{\theta_{old}}(o_i|q)}A_i,\operatorname{clip}\left(\frac{\pi_\theta(o_i|q)}{\pi_{\theta_{old}}(o_i|q)},1-\varepsilon,1+\varepsilon\right)A_i\right)-\beta\mathbb{D}_{KL}\left(\pi_\theta||\pi_{ref}\right)\right)
\end{align}
where $\varepsilon$ and $\beta$ are hyperparameters, $A_i$ is the advantage derived from relative rewards within each group, and $\pi_{ref}$ is the reference model for computing KL divergence. The core idea of GRPO is to evaluate the relative quality of multiple candidate responses within a group. For a question $q$, GRPO generates $G$ distinct responses $\{o_1, o_2, \dots, o_G\}$ using $\pi_{\theta_{old}}$, computes their rewards $\{r_1, r_2, \dots, r_G\}$ via a reward function, and calculates normalized scores to simulate PPO's advantage function:
\begin{align}
A_i=\frac{r_i-\mathrm{mean}(\{r_1,\ldots,r_G\})}{\mathrm{std}(\{r_1,\ldots,r_G\})}
\end{align}
This approach not only reduces reliance on external critic models but also enhances the model's ability to distinguish high-quality from low-quality outputs, mitigating hardware constraints for reasoning in generating long CoT for self-reflection and correction. Building on this, DeepSeek R1 employs verifiable rewards to optimize task models for objectives with objectively verifiable results, eliminating the need for intermediate reward modeling. Examples include mathematical problem-solving, coding challenges, and domains with well-defined correctness criteria. This method addresses reward hacking and establishes a foundational paradigm for RL-based reasoning. 

\subsubsection{Reward Mechanisms}

To effectively guide model behavior during inference, particularly in complex reasoning tasks, two types of reward mechanisms are commonly used: Outcome Reward Mechanisms (ORM), which focus on the correctness of final outputs, and Process Reward Mechanisms (PRM), which emphasize the quality of intermediate reasoning steps. These mechanisms offer distinct advantages and pose different challenges in model training and alignment.

\paragraph{Outcome Reward Mechanisms} 
ORM primarily focus on evaluating an agent based on the final outcome or result achieved after executing a sequence of actions. These mechanisms typically assess policy performance according to the terminal state of the task or the total accumulated reward throughout the process. The reward signals in such settings are generally sparse and delayed, providing feedback only upon task completion or successful attainment of the final goal. In the context of model reasoning tasks, ORM refers to assessing and optimizing the model based on the correctness of its final output. The reward is thus conditioned solely on the correctness of the answer, without regard for the model’s intermediate reasoning behavior. For instance, in mathematical problem-solving or question answering tasks, the model receives a positive reward only when the final answer is correct \cite{shao2024deepseekmath, guo2025deepseek}.
Due to its simplicity and the absence of complex reward model training, ORM has become a widely adopted strategy for RL-based reasoning in LLMs and MLLMs. However, ORM inherently suffers from the temporal credit assignment problem, where the agent struggles to discern which specific actions contributed positively or negatively to the final outcome \cite{sutton1998reinforcement}. This issue is particularly pronounced in environments with long temporal horizons or high-dimensional action spaces, where it significantly impedes learning efficiency. Furthermore, the sparse and delayed nature of the reward signal can lead to low sample efficiency and unstable gradient updates during training.

\paragraph{Process Reward Mechanisms} 
In contrast to ORM, PRM place emphasis on evaluating the model’s intermediate behaviors during the reasoning process, thereby encouraging the incremental formation of coherent and logically sound CoT. Rather than focusing solely on task completion, PRM rewards intermediate behaviors that align with desirable reasoning patterns, even in cases where the final objective is not fully achieved. In model inference tasks, this approach is commonly employed to promote the generation of logically consistent intermediate reasoning steps, plausible deductive structures, or the use of credible sources from external knowledge bases. Recent research has proposed the incorporation of process-level reward signals—such as logical consistency, informational completeness, or citation reliability for each reasoning step—into the fine-tuning process, thereby improving model interpretability and enhancing user trust \cite{zhang2025r1, fan2025chestx}. Compared to outcome-based rewards, process-level mechanisms offer finer-grained supervision, facilitating more effective debugging and alignment of model behavior. Nevertheless, the design of process rewards depends heavily on the accurate evaluation of intermediate reasoning steps, which poses significant challenges for automation. Moreover, the absence of standardized and robust evaluation criteria limits the scalability and generalizability of PRMs across diverse tasks and model architectures.

\subsubsection{Training Efficiency}

Improving training efficiency has become a critical focus in RL, particularly for real-world applications where computational resources and data are limited. Two prominent strategies for enhancing training efficiency are curriculum reinforcement learning and data-efficient learning techniques.

Curriculum reinforcement learning draws inspiration from the human learning process by introducing tasks in a structured progression, starting from easier subtasks and gradually moving to more complex ones \cite{deng2025boosting, meng2025mm}. This staged training paradigm allows agents to accumulate knowledge incrementally, leading to faster convergence and improved performance on challenging tasks. It is particularly effective in environments with a clear task difficulty hierarchy.

Data-efficient learning aims to maximize performance with minimal data. Two commonly adopted approaches in this category include: (1) Prioritized Sampling \cite{team2025kimi}: Rather than uniformly sampling from experience replay buffers, prioritized sampling selects transitions based on their learning potential. This strategy focuses updates on informative samples, accelerating learning and improving policy quality. (2) High-Quality Samples \cite{kang2025gflowvlm}: Leveraging samples with higher utility—such as trajectories with higher rewards or successful episodes—can significantly improve sample efficiency. By concentrating training on more informative experiences, the agent learns more effectively while reducing unnecessary computational overhead.

\subsection{MLLM Reasoning}

\subsubsection{Differences and Similarities of Reasoning in LLMs and MLLMs}

\begin{table}[t!]
\begin{tabular}{lll}
\toprule
Aspect           & LLMs             & MLLMs                \\ \midrule
Architecture     & Transformer (text-only)        & Transformer (multimodal)             \\
CoT Structure    & Text-based reasoning           & Cross-modal reasoning \\
Input Modality   & Text-only                      & Text, image, audio, video, etc              \\
Reasoning        & Linear textual steps           & Integrated reasoning across multiple modalities    \\
Grounding        & Language-based knowledge       & Visual, textual, or other sensory data         \\
Applications     & Text-based tasks & Multimodal tasks    \\
Training Data    & Text corpora                   & Multimodal datasets                  \\
Interpretability & Textual reasoning steps        & Multimodal reasoning steps           \\
Complexity       & Simpler (text-only)            & Higher complexity (multimodal fusion)  \\ \bottomrule
\end{tabular}
\caption{Differences in reasoning  between LLMs and MLLMs}
\label{tab:llmvsmllm}
\end{table}

LLMs and MLLMs both employ CoT-based reasoning to address complex tasks, with the core principle being the decomposition of intricate problems into intermediate steps that guide the model toward a solution. Whether prompted through textual examples for LLMs or multimodal inputs for MLLMs, the CoT process remains fundamentally similar: the model systematically builds its reasoning based on prior steps, providing transparency into its internal processes. This transparency is particularly valuable for tasks like mathematical reasoning and code generation, where intermediate steps are crucial for interpreting outputs. However, significant differences arise due to the types of inputs processed. We summarize the main differences in CoT-based reasoning between LLMs and MLLMs in Table \ref{tab:llmvsmllm}. In LLMs, CoT reasoning is confined to linguistic operations, involving logical deduction, arithmetic computation, or language clarification within a linear, text-centric framework. Conversely, MLLMs engage in more dynamic and nonlinear reasoning by alternating between textual and other modalities, such as visual perception. For instance, an MLLM might first reason about visual input such as identifying "a blue car" in an image, and then integrate this with textual reasoning, such as referencing typical car colors. This integration of modalities makes MLLMs' CoT reasoning inherently multidimensional but introduces challenges like cross-modal alignment, accurate incorporation of visual features, and avoiding modality bias. Grounding further differentiates the two, as MLLMs anchor reasoning in real-world knowledge, such as visual objects or spatial relationships, while LLMs rely solely on textual and abstract concepts. In summary, while both models leverage CoT reasoning, LLMs excel in text-centered reasoning, whereas MLLMs expand this capability by integrating multimodal data, necessitating tailored approaches that account for their unique characteristics and enable nonlinear interactions across modalities in MM-CoT reasoning.

\subsubsection{Methodologies in MLLM Reasoning}

Inspired by the significant potential of reasoning in NLP for solving complex language tasks \cite{guo2025deepseek, jaech2024openai}, recent studies have sought to enhance the reasoning capabilities of MLLMs. Broadly, current MLLM reasoning methods fall into two main categories:

\paragraph{SFT-Based} The first method focuses on guiding models to generate high-quality CoT data that adheres to specific steps. This can be achieved either through structured reasoning templates—such as summaries, descriptions, logical reasoning, and conclusions—or by leveraging advanced search algorithms. Notable examples include CoT prompting techniques \cite{zhao2023antgpt, fei2024video}, planning-based approaches like Best-of-N \cite{chen2024alphamath, xu2024llava}, Monte Carlo Tree Search (MCTS) \cite{yao2024mulberry, xie2024monte}, and Beam Search \cite{wu2024v, yao2023tree}. The generated CoT data is then utilized for SFT, enabling MLLMs to emulate high-quality, structured reasoning. While this approach ensures the quality of CoT data and enables effective integration of multi-modal information, it suffers from two key limitations: the labor-intensive nature of CoT annotation and the risk of catastrophic forgetting during fine-tuning.

\paragraph{RL-Based} The second category builds on the DeepSeek R1 \cite{guo2025deepseek} reinforcement learning paradigm, which integrates large-scale RL with structured and outcome-driven reward functions. This framework enables large language models to automatically generate human-like reasoning processes with remarkable complexity \cite{team2025kimi, liu2025visual, zhao2025r1, meng2025mm, peng2025lmm}, eliminating the need for labor-intensive CoT annotations. Additionally, it demonstrates strong generalization capabilities when handling OOD data.However, current RL-based reasoning methods for MLLMs are primarily adapted from NLP, with the generated reasoning trajectories predominantly presented in textual form, neglecting interactions among diverse modalities such as images, videos, and audio in MM-CoT. Additionally, the lack of effective reasoning path rewards in current RL-based reasoning often leads to CoT outputs containing excessive redundant tokens. 

\subsubsection{SFT Memory, RL Generalization}
SFT-based reasoning methods activate the model's inherent reasoning capabilities by identifying precise cognitive templates and directly constructing high-quality reasoning trajectories. In contrast, RL-based approaches focus on leveraging RL to search for optimal reasoning paths among numerous possibilities, treating the cultivation of reasoning abilities as a process that requires extensive training to explore effective reasoning patterns. This distinction is now widely recognized by the principle: SFT for memory, RL for generalization \cite{chu2025sft}. SFT offers strong inductive biases and precise knowledge injection, whereas RL unleashes the model’s potential for abstract, adaptive, and robust reasoning under uncertainty. In the context of MLLMs, where cross-modal alignment and dynamic understanding are critical, this dichotomy becomes even more pronounced. Therefore, advancing RL-based reasoning adaptation techniques tailored for multimodal scenarios is essential. This study contributes to this direction by providing a comprehensive review and analysis of RL-based reasoning strategies in MLLMs.

\section{Key Design and Optimization of RL Algorithms in LLMs/MLLMs}
\label{sec:RL}

In RL for LLMs, methods can be classified into value-model-free such as GRPO \cite{shao2024deepseekmath} and value-model-based such as PPO \cite{schulman2017proximal} approaches, depending on whether a value model is learned \cite{yuan2025vapo}. While GRPO exhibits entropy collapse and instability, PPO faces challenges in long-chain reasoning. Recent optimizations address these limitations, demonstrating improved performance in LLMs, with potential applicability to MLLMs for enhanced training efficiency and stability.

\subsection{Value-Model-Free}

Value-model-free RL methods, such as GRPO \cite{shao2024deepseekmath}, have demonstrated remarkable effectiveness in LLM training by eliminating the need for value model computation. Instead, these approaches derive advantages from trajectory-level rewards uniformly assigned to each token, providing stability through group-based reward averaging. This characteristic proves particularly valuable in scenarios where training value models is challenging, making value-model-free methods highly successful in complex tasks like long-chain reasoning. While these methods offer implementation simplicity and robustness, they also reveal fundamental limitations, such as entropy collapse and reward noise, which recent advancements aim to address through novel optimization techniques.

\subsubsection{DAPO}

Dynamic sAmpling Policy Optimization (DAPO) \cite{yu2025dapo} significantly improves the model's generation diversity and RL training efficiency by introducing an asymmetric clipping strategy, a dynamic sampling mechanism, a token-level policy gradient loss, and extended reward shaping. The optimization formula is as follows:
\begin{align}
\label{equ:DAPO}
    J_{\mathrm{DAPO}}(\theta)&=\mathbb{E}_{(q,a)\sim D,\{o_i\}_{i=1}^G\sim\pi_{\theta_{\mathrm{old}}}{(\cdot|q)}} \notag \\  &\left[\frac{1}{\sum_{i=1}^G\|o_i\|}\sum_{i=1}^G\sum_{t=1}^{\|o_i\|}\min\left(r_{i,t}(\theta)\hat{A}_{i,t},\operatorname{clip}(r_{i,t}(\theta),1-\epsilon_{\mathrm{low}},1+\epsilon_{\mathrm{high}})\hat{A}_{i,t}\right)\right] \\
    \mathrm{s.t.} \quad &0<\left|\{o_i\mid\mathtt{is\_equivalent}(a,o_i)\}\right|<G, \notag
\end{align}
where $\hat{A}_{i,t}=\frac{r_i-\text{mean}(\{R_i\}_{i=1}^G)}{\mathrm{std}(\{R_i\}_{i=1}^G)}$, and the other symbols are consistent with those in Equation \ref{equ:GRPO}.

\paragraph{Clip-Higher Strategy}
In GRPO, the clipping range is typically symmetric, i.e., $\epsilon_{\mathrm{low}} = \epsilon_{\mathrm{high}}$. However, this symmetric clipping strategy limits the exploration of low-probability tokens, potentially leading to entropy collapse. To address this issue, DAPO introduces the Clip-Higher strategy in Equation \ref{equ:DAPO}, decoupling the clipping bounds into $\epsilon_{\mathrm{low}}$ and $\epsilon_{\mathrm{high}}$. This asymmetric clipping strategy significantly increases the value of $\epsilon_{\mathrm{high}}$, thereby allowing for greater flexibility in enhancing low-probability tokens.

\paragraph{Dynamic Sampling}
To address the issue where gradient signals vanish if all samples in a batch have accuracy rates of either 1 or 0, DAPO proposes a dynamic sampling strategy. By filtering out samples with accuracy rates of 0 or 1, this strategy ensures that every batch contains samples with valid gradient signals, represented as:
$0 < |\{o_i|\text{is equivalent}(a,o_i)\}| < G.$
The authors demonstrate that dynamic sampling can substantially improve training efficiency.

\paragraph{Token-Level Policy Gradient Loss}
GRPO computes loss at the sample level, averaging losses across tokens within each sample before aggregating across different samples. This approach poses challenges in long-chain reasoning scenarios, as the contribution of tokens in longer sequences to the overall loss is diluted, making it difficult for the model to learn high-quality reasoning patterns. To address this, DAPO proposes token-level policy gradient loss, which ensures that tokens in longer sequences contribute gradients more fairly.

\paragraph{Overlong Reward Shaping}
To address reward noise  caused by improper shaping of truncated samples, the Overlong Reward Shaping strategy introduces a progressive length-dependent penalty mechanism. This adjusts rewards for truncated outputs and discourages excessive verbosity by applying scaled punishments proportional to response lengths beyond a threshold, as follows:
\begin{align}
    R_\mathrm{length}(y)=
\begin{cases}
0, & \|y\|\leq L_\mathrm{max}-L_\mathrm{cache}, \\
\frac{(L_\mathrm{max}-L_\mathrm{cache})-\|y\|}{L_\mathrm{cache}}, & L_\mathrm{max}-L_\mathrm{cache}<\|y\|\leq L_\mathrm{max}, \\
-1, & \|y\|>L_\mathrm{max}.
\end{cases}
\end{align}
Where $L_\mathrm{cache}$ is the buffer length, and $L_\mathrm{max}$ is the maximum allowable output length for the model. This mechanism significantly reduces the impact of reward noise, stabilizing the training process.

\subsubsection{Dr.GRPO}

Dr.GRPO \cite{liu2025understanding} highlights that GRPO exhibits two biases: a response length bias and a problem difficulty bias. Specifically, GRPO introduces normalization terms for output length and reward standard deviation when computing the advantage function. These terms cause the model to favor generating longer incorrect responses during optimization and assign different weights to problems of varying difficulty. To address these issues, the proposed method eliminates these biases by removing the normalization terms. The objective function of Dr.GRPO is as follows:
\begin{align}
\label{qua:dr.grpo}
    J_{\text{Dr.GRPO}}(\pi_\theta) &= \mathbb{E}_{q \sim p_Q, \{o_i\}_{i=1}^G \sim \pi_{\theta_{\text{old}}}(\cdot|q)} \\ \notag &\left[ \frac{1}{G} \sum_{i=1}^G \sum_{t=1}^{\|o_i\|} \min \left( \frac{\pi_\theta(o_{i,t}|q, o_{i,<t})}{\pi_{\theta_{\text{old}}}(o_{i,t}|q, o_{i,<t})} \tilde{A}_{i,t}, \text{clip} \left( \frac{\pi_\theta(o_{i,t}|q, o_{i,<t})}{\pi_{\theta_{\text{old}}}(o_{i,t}|q, o_{i,<t})}, 1-\epsilon, 1+\epsilon \right) \tilde{A}_{i,t} \right) \right],
\end{align}
where the advantage function is defined as: $\tilde{A}_{i,t} = R(q, o_i) - \text{mean}(\{R(q, o_1), \dots, R(q, o_G)\})$.

\paragraph{Removal of Length Normalization}
Compared to Equation \ref{equ:GRPO}, the authors remove the normalization term $\|o_i\|$, which avoids the preference for longer incorrect responses and prevents the model from generating increasingly lengthy errors.
  
\paragraph{Removal of Standard Deviation Normalization}
For the advantage in Equation \ref{equ:GRPO}, the authors eliminate the normalization term $\text{std}(\{R(q, o_1), \dots, R(q, o_G)\})$. This removes the problem difficulty bias, ensuring that problems of different difficulties are weighted equally during optimization. 

\subsubsection{CPPO}

Completion Pruning Policy Optimization (CPPO) \cite{lin2025cppo} is a novel approach designed to accelerate the training of inference models based on GRPO. Compared to GRPO, CPPO introduces several improvements, primarily focusing on reducing computational overhead, enhancing training efficiency, and maintaining or even improving model performance. The objective function of CPPO builds upon GRPO by incorporating pruning conditions, defined as follows:
\begin{equation}
\begin{split}
J_{CPPO}(\theta) = & \mathbb{E}_{q \sim P(Q), \{o_i\}_{i=1}^G \sim \pi_{\theta_0 Id}(o|q)} \Bigg\{ \frac{1}{k} \sum_{i \in I} \frac{1}{|o_i|} \sum_{t=1}^{|o_i|} \min \Bigg[ \frac{\pi_\theta(o_{i,t}|q, o_{i,<t})}{\pi_{\theta_0 Id}(o_{i,t}|q, o_{i,<t})} A_i, \\
&\text{clip}\left( \frac{\pi_\theta(o_{i,t}|q, o_{i,<t})}{\pi_{\theta_0 Id}(o_{i,t}|q, o_{i,<t})}, 1 - \epsilon, 1 + \epsilon \right) A_i - \beta \mathbb{D}_{KL} \left[ \pi_\theta \|\pi_{ref} \right] \Bigg) \Bigg\},
\end{split}
\end{equation}
where $\mathcal{I} = \{i \in \{1, \dots, G\} \mid |A_i| \text{ is among the top } k \text{ values}\}$, meaning only the top $k$ completions with the highest absolute advantage values are retained for gradient updates.

\paragraph{Strategy to Reduce Computational Overhead} CPPO identifies that the contribution of completions to training is closely related to their absolute advantage values. Completions with low advantage values contribute minimally to the training signal and may even introduce noise. Therefore, in Equation (5), CPPO retains only the top $k$ completions with the highest absolute advantage values for gradient updates, pruning the others to reduce redundant computations.

\paragraph{Dynamic Completion Allocation Strategy} While the completion pruning strategy reduces computational overhead, it causes underutilization of GPU resources. After pruning, the reduced number of retained completions fails to fully leverage the parallel computing capabilities of GPUs. To address this, CPPO combines the remaining pruned completions with high-quality completions from new queries, filling the GPU processing pipeline. This ensures that each GPU device operates at full capacity, thereby maximizing the utilization of GPU parallel computing power. 

\subsubsection{GPG} 

Group Policy Gradient (GPG) \cite{chu2025gpg} eliminates the reliance on a reference model and avoids KL divergence constraints by directly optimizing the original RL objective, thereby simplifying the training process and reducing computational costs. Additionally, GPG introduces an Accurate Gradient Estimation (AGE) technique and a group reward normalization method, effectively addressing issues related to advantage function and gradient estimation bias. These innovations enhance training stability and performance.

\paragraph{Direct Optimization of the Original RL Objective.} 
Methods like GRPO typically rely on surrogate loss functions to approximate policy optimization. By contrast, GPG directly optimizes the original RL objective, avoiding biases that surrogate loss functions might introduce. Its core optimization objective is formulated as follows:  
\begin{align}
\label{equ:gpg}
    J_{\mathrm{GPG}}(\theta)=\mathbb{E}_{(q,a)\sim D,\{o_i\}_{i=1}^G}\left[\frac{1}{\sum_{i=1}^G\|o_i\|}\sum_{i=1}^G\sum_{t=1}^{\|o_i\|}\left(-\log\pi_\theta(o_{i,t}|q,o_{i,<t})A_{i,t}\right)\right],
\end{align}
where $ A_{i,t} $ is the normalized intra-group advantage function.

\paragraph{Elimination of the Reference Model}
As shown in Equation \ref{equ:GRPO}, GRPO stabilizes training by using a reference model and KL divergence constraints, whereas in Equation \ref{equ:gpg} GPG completely removes these components. This results in a more concise loss function for GPG.

\paragraph{Addressing Advantage Function and Gradient Estimation Bias}
GPG proposes an Accurate Gradient Estimation (AGE) technique, which effectively resolves the issue of reward bias, formulated as:  
\begin{align}
    \hat{g}=\frac{\sum_{i=M+1}^Bg_i}{B-M}=g\cdot\alpha,\quad\alpha=\frac{B}{B-M},
\end{align}
where $ M $ represents the number of invalid samples (i.e., samples with zero gradients), and $ \alpha $ is a dynamic adjustment factor used to correct gradient estimation bias.  

\subsection{Value-Model-Based}

Value-model-based methods, exemplified by PPO,excel by enabling precise, step-wise credit assignment—an especially crucial feature for complex reasoning tasks, where minor errors at individual steps can propagate into catastrophic failures. In contrast to the advantage estimates from Monte Carlo methods in value-model-free approaches, a well-trained value model not only provides lower-variance token-level value estimates for enhanced training stability, but also exhibits inherent generalization capabilities that improve online sample utilization efficiency\cite{yuan2025vapo}. While training value models for long CoT reasoning remains challenging, recent studies have introduced innovative optimization techniques that significantly improve both stability and overall performance.

\subsubsection{ORZ}
Open-Reasoner-Zero (ORZ) \cite{hu2025open} demonstrates through extensive experiments that by using the vanilla PPO algorithm combined with GAE ($\lambda=1$, $\gamma=1$) and a simple rule-based reward function, significant improvements in response length and benchmark performance can be achieved without any KL regularization. Additionally, the authors highlight the importance of increasing both the quantity and diversity of training data for enhancing model performance. Compared to DeepSeek-R1-Zero, ORZ achieves comparable performance with only one-tenth of the training steps.

\subsubsection{VC-PPO}
 
To enable traditional reinforcement learning algorithms like PPO to handle long CoT tasks, Value-Calibrated PPO (VC-PPO) \cite{yuan2025s} introduces optimizations in two key areas: value initialization bias and Decoupled-GAE.  

\paragraph{Value Initialization Bias}
The value function $V(s)$ in PPO is typically initialized using a reward model during the early stages of training. However, this initialization method can lead to significant bias, especially in long-sequence tasks. In the initial phase, since the reward model tends to underestimate the scores of earlier tokens, the advantage function $A_t$ exhibits a positive bias when estimating the contributions of these tokens. This bias encourages the policy to favor shorter outputs, thereby undermining the model's ability to generate long CoT sequences. To address this issue, VC-PPO proposes \textit{Value Pretraining}. By offline training the value model under a fixed policy and using GAE with $\lambda = 1.0$ to ensure unbiased returns, the value model achieves a more accurate estimation of expected rewards.

\paragraph{Decoupled-GAE}  
In long-sequence tasks, reward signals are often concentrated at the end of the sequence. While conventional RL tasks typically employ GAE ($\lambda=0.95$) to mitigate variance in cumulative rewards, RLHF fundamentally differs due to its trajectory-level reward signals. When $\lambda < 1.0$, these reward signals decay rapidly as they propagate backward through the sequence, which significantly hampers the learning effectiveness of the value model. Decoupled-GAE addresses the distinct variance reduction needs in RLHF by decoupling $\lambda$ values for policy and value optimization.  Here, $\lambda<1.0$ induces severe signal attenuation across long trajectories, making unbiased value estimation ($\lambda=1.0$) essential while retaining $\lambda=0.95$ for stable policy updates. This dual-$\lambda$ mechanism independently optimizes the bias-variance trade-off for each component, achieving superior training efficiency over standard GAE methods, the  formula is expressed as follows:  
\begin{align}
    V_{\text{target}}(s_t) = 
  \begin{cases} 
  \sum_{l=0}^{T-t-1} \lambda^l (r_{t+l} + V(s_{t+l+1}) - V(s_{t+l})) + V(s_t), & \lambda < 1.0 \\
  \sum_{l=0}^{T-t-1} r_{t+l}, & \lambda = 1.0
  \end{cases}
\end{align}
With this design, the value model estimates expected rewards with reduced bias, while the policy model accelerates convergence by reducing variance.

\subsubsection{VAPO} 
Value-model-based Augmented Proximal Policy Optimization (VAPO) \cite{yuan2025vapo} enhances the training stability of PPO in long-chain CoT reasoning tasks. It adopts several tricks from DAPO \cite{yu2025dapo} and VC-PPO \cite{yuan2025s}, such as the Clip-Higher Strategy, Token-Level Policy Gradient Loss, and Decoupled-GAE, while introducing Length-Adaptive GAE, Positive Example LM Loss, and Group-Sampling to achieve balanced optimization across varying sequence lengths and mitigate the issue of sparse rewards.

\paragraph{Length-Adaptive GAE} 
Building on the Decoupled-GAE method in VC-PPO, VAPO introduces a Length-Adaptive GAE approach that dynamically adjusts the $\lambda$ parameter to balance the bias-variance tradeoff for sequences of varying lengths. The specific formula is as follows:
\begin{align}
    \lambda_{\text{policy}} = 1 - \frac{1}{\alpha l},
\end{align}
where $ l $ denotes the sequence length and $ \alpha $ is a hyperparameter that controls the overall bias-variance trade-off. This formula ensures a more uniform distribution of the TD-error coefficients, preventing the excessive decay of reward signals in long sequences. 

\paragraph{Positive Example LM Loss}
VAPO incorporates the Positive Example LM Loss, which leverages the negative log-likelihood (NLL) of the correct answers to improve the utilization efficiency of positive samples. The corresponding formula is:
\begin{align}
    \mathcal{L}_{\text{NLL}}(\theta) = - \frac{1}{\sum_{o_i \in T} \|o_i\|} \sum_{o_i \in T} \sum_{t=1}^{\|o_i\|} \log \pi_{\theta}(a_t \mid s_t),
\end{align}
where $ T $ denotes the set of correct answers, and $ \|o_i\| $ represents the length of the sequence $ o_i $. 

\paragraph{Group-Sampling}
VAPO employs Group-Sampling to extract discriminative positive and negative samples within the same prompt. It highlights that, given a fixed computational budget, reducing the number of distinct prompts per batch and reallocating resources for repeated generations improves performance. This is attributed to the introduction of richer contrastive signals, which enhances the learning capability of the policy model.

\section{RL-Based MLLM Reasoning}
\label{sec:MLLMReasoning}

\begin{table}[t!]
\centering
\resizebox{\textwidth}{!}{
\begin{tabular}{lccccc}
\toprule
\multicolumn{1}{l|}{Model}                                                               & Base Models& Strategy & Algorithm  & Applications        & MM-CoT    \\ \midrule
\multicolumn{6}{c}{Image-Based}\\ \midrule
\multicolumn{1}{l|}{KIMI K1.5 \cite{team2025kimi}}                      & KIMI K1.5& SFT\&RL  & OPMD       & Univ. Reason     & T         \\
\multicolumn{1}{l|}{MedVLM-R1 \cite{pan2025medvlm}}                     & Qwen2-VL-2B& RL       & GRPO       & Medical             & T         \\
\multicolumn{1}{l|}{Visual-RFT \cite{liu2025visual}}                    & Qwen2-VL-2B& RL       & GRPO       & Detection\&CLS      & T         \\
\multicolumn{1}{l|}{VisualThinker-R1-Zero \cite{zhou2025r1}}            & Qwen2-VL-2B& RL       & GRPO       & Spatial Reason   & T         \\
\multicolumn{1}{l|}{Vision-R1 \cite{huang2025vision}}                   & Qwen-2.5-VL-7B-Instruct& SFT\&RL  & GRPO\&PTST & Univ. Reason     & T         \\
\multicolumn{1}{l|}{Vision-R1 \cite{zhan2025vision}}                   & Qwen2.5-VL-7B, Griffon-G-7B \cite{zhan2024griffon}& SFT\&RL  & GRPO & Detection     & T         \\
\multicolumn{1}{l|}{Seg-Zero \cite{liu2025seg}}                         & Qwen2.5-VL-3B& RL       & GRPO       & Segmentation        & T         \\
\multicolumn{1}{l|}{GFlowVLM \cite{kang2025gflowvlm}}                   & LLaVA-v1.6-Mistral-7B& SFT\&RL  & GFlowNet   & Univ. Reason     & T         \\
\multicolumn{1}{l|}{MM-Eureka \cite{meng2025mm}}                     & InternVL2.5-Instruct-8/38B& RL       & RLOO       & Math                & T         \\
\multicolumn{1}{l|}{Curr-ReFT \cite{deng2025boosting}}                  & Qwen2.5-VL-3/7B& RL       & GRPO       & Univ. Reason     & T         \\
\multicolumn{1}{l|}{LMM-R1 \cite{peng2025lmm}}                          & Qwen2.5-VL-Instruct-3B& RL       & PPO        & Univ. Reason     & T         \\
\multicolumn{1}{l|}{R1-Onevision \cite{yang2025r1}}                     & Qwen2.5-VL-3/7B& SFT\&RL  & GPRO       & Univ. Reason     & T\&I Desc \\
\multicolumn{1}{l|}{R1-VL \cite{zhang2025r1}}                           & Qwen2.5-VL-7B& SFT\&RL  & StepGRPO   & Univ. Reason     & T\&I Desc \\
\multicolumn{1}{l|}{Skywork R1V \cite{skywork2025r1v}}                  & \makecell{ DeepSeek-R1-Distill-Qwen-32B/ \\ QwQ-32B\&InternViT-6B-448px-V2\_5} & SFT\&RL  & GRPO       & Univ. Reason     & T         \\
\multicolumn{1}{l|}{OThink-MR1 \cite{liu2025othink}}                    & Qwen2-VL-Instruct-2/7B& RL       & GRPO-D     & Univ. Reason     & T         \\
\multicolumn{1}{l|}{OpenVLThinker \cite{deng2025openvlthinker}}      & Qwen2.5-VL-Instruct-7B& SFT\&RL  & GRPO       & Math                & T         \\
\multicolumn{1}{l|}{MetaSpatial \cite{pan2025metaspatial}}              & Qwen-VL-3B/7B& RL       & GRPO       & Univ. Reason     & T         \\
\multicolumn{1}{l|}{UI-R1 \cite{lu2025ui}}                              & Qwen2.5-VL-3B& RL       & GRPO       & GUI   & T         \\
\multicolumn{1}{l|}{Reason-RFT \cite{tan2025reason}}                              & Qwen2-VL-Instruct-3/7B & RL       & GRPO       & Univ. Reason   & T         \\ 
\multicolumn{1}{l|}{Q-Insight \cite{li2025q}}                              & Qwen-2.5-VL-7B & RL       & GRPO       & Image Quality   & T\&I Desc  \\ 
\multicolumn{1}{l|}{Kimi-VL \cite{team2025kimivl}}                              & Kimi-VL & SFT\&RL  & OPMD       & Univ. Reason   & T  \\
\multicolumn{1}{l|}{ThinkLite-VL \cite{wang2025sota}}                              & Qwen2.5-VL-Instruct-7B & RL  & GRPO       & Univ. Reason   & T  \\
\multicolumn{1}{l|}{Perception-R1 \cite{yu2025perception}}                              & Qwen2-VL-Instruct-2B & RL  & GRPO       & Univ. Reason   & T  \\
\multicolumn{1}{l|}{VLAA-Thinker \cite{chen2025sft}}                              & Qwen2-VL-2/7B, Qwen2.5-VL-3/7B & RL  & GRPO       & Univ. Reason   & T  \\
\multicolumn{1}{l|}{VL-Rethinker \cite{wang2025vl}}                              & Qwen2.5-VL-7/72B & RL  & GRPO-SSR    & Univ. Reason   & T  \\
\multicolumn{1}{l|}{VLM-R1 \cite{shen2025vlm}}                              & Qwen2.5-VL-Instruct-3/7/32B & RL  & GRPO    & Univ. Reason   & T  \\
\multicolumn{1}{l|}{GUI-R1 \cite{xia2025gui}}                              & Qwen2.5-VL-3/7B & SFT\&RL  & GRPO    & GUI   & T  \\
\multicolumn{1}{l|}{InfiGUI-R1 \cite{liu2025infigui}}                              & Qwen2.5-VL-3B-Instruct & SFT\&RL  & RLOO    & GUI   & T  \\
\multicolumn{1}{l|}{NoisyRollout \cite{liu2025noisyrollout}}                              & Qwen2.5-VL-7B-Instruct & RL  & GRPO    & Math   & T  \\
\multicolumn{1}{l|}{R1-SGG \cite{chen2025compile}}                              & Qwen2-VL-2/7B & SFT\&RL  & GRPO    & Generation   & T  \\
\multicolumn{1}{l|}{Relation-R1 \cite{li2025relation}}                              & Qwen2.5-VL-3B & SFT\&RL  & GRPO    & Spatial Reason   & T  \\
\multicolumn{1}{l|}{Skywork R1V2 \cite{wei2025skywork}}                              & InternViT-6B, QwQ-32B & MPO\&RL  & GRPO-SSB    & Univ. Reason   & T  \\
\multicolumn{1}{l|}{FAST \cite{xiao2025fast}}                              & Qwen2.5-VL-3/7B & RL  & Fast-GRPO    & Univ. Reason   & T  \\
\multicolumn{1}{l|}{ChestX-Reasoner \cite{fan2025chestx}}                              & Qwen2-VL-7B & SFT\&RL  & GRPO    & Medical   & T  \\
\midrule
\multicolumn{6}{c}{Video-Based} \\ \midrule
\multicolumn{1}{l|}{Video-R1 \cite{feng2025video}}                      & Qwen2.5-VL-7B& SFT\&RL  & T-GRPO     & Univ. Reason     & T         \\ 
\multicolumn{1}{l|}{R1-Zero-VSI \cite{liao2025improved}}                              & Qwen2-VL-2/72B & RL  & GRPO    & Spatial Reason   & T\&I Desc  \\
\multicolumn{1}{l|}{Spatial-R1 \cite{ouyang2025spatial}}                              & Qwen2.5-VL-7B-Instruct & RL  & GRPO    & Spatial Reason   & T  \\
\multicolumn{1}{l|}{Embodied-R \cite{zhao2025embodied}}                              & \makecell{ Qwen2.5-3B-Instruct \\ Qwen2.5-VL-72B-Instruct} & RL  & GRPO    & Spatial Reason   & T  \\
\multicolumn{1}{l|}{TinyLLaVA-Video-R1 \cite{zhang2025tinyllava}}                              & \makecell{ Qwen2.5-3B\&SigLIP} & RL  & GRPO    & ST-Perception   & T  \\
\multicolumn{1}{l|}{VideoChat-R1 \cite{li2025videochat}}                              & \makecell{ Qwen2/2.5-VL-7B} & RL  & GRPO    & ST-Perception   & T  \\
\midrule
\multicolumn{6}{c}{Audio-Based} \\ \midrule
\multicolumn{1}{l|}{R1-AQA 
\cite{li2025reinforcement}} & Qwen2-Audio-7B-Instruct& RL       & GRPO       & Audio QA            & T         \\
\multicolumn{1}{l|}{SARI
\cite{wen2025sari}} & \makecell{ Qwen2-Audio-7B-Instruct \\ Qwen2.5-Omni }& SFT\&RL       & GRPO       & Audio QA            & T         \\
\midrule
\multicolumn{6}{c}{Omni} \\ \midrule
\multicolumn{1}{l|}{R1-Omni \cite{zhao2025r1}}                          & HumanOmni-0.5B& SFT\&RL       & GRPO       & Emotion & T \\ \bottomrule 
\end{tabular}}
\caption{Overview of multimodal reasoning models employing reinforcement learning strategies. \textit{Univ. Reason} denotes general-purpose reasoning across diverse tasks, \textit{Image Quality} refers to the image quality understanding task, \textit{Emotion} represents the emotion  recognition task and \textit{ST-Perception} refers to the spatio-temporal perception tasks. In the MM-CoT column, \textit{T} indicates support for the text modality, \textit{I} for the image modality.
}
\label{tab:Multi-Model-Reasoning-Survey}
\end{table}

Deepseek-R1 \cite{guo2025deepseek} demonstrates that the reasoning capabilities of LLMs can be activated through pure RL. Current RL-based reasoning methods model the reasoning process as a Markov Decision Process (MDP). The core objective is to enhance the model's reasoning capability by maximizing the expected reward of the reasoning path $ S $. Recently, researchers have begun exploring how to apply the R1 training paradigm to MLLMs to further improve their reasoning performance in complex tasks. In Table \ref{tab:Multi-Model-Reasoning-Survey}, we summarize the current research progress of reasoning methods based on RL in the multimodal domain.
In Section \ref{subsec:simple-RL}, we review recent efforts to adapt the R1 training paradigm from LLMs to MLLMs, highlighting key challenges and corresponding solutions. Section \ref{subsec:rm4mllm} categorizes reward mechanisms in multimodal settings into Outcome-supervised and Process-supervised types. Section \ref{subsec:effAndStab} discusses methods that enhance training efficiency and stability through data utilization. Finally, Section \ref{subsec:RLHF-Framwork} summarizes prevalent RLHF frameworks that support rapid iteration of RL-based reasoning approaches.

\subsection{RL Training Paradim: From LLM to MLLM}
\label{subsec:simple-RL}
\subsubsection{Standardized R1 Training Paradigm in LLM}
The success of Kimi K1.5 \cite{team2025kimi} and DeepSeek R1 \cite{guo2025deepseek} in enhancing model reasoning capabilities through RL has driven the rise of the R1 training paradigm in the field of model reasoning.

\paragraph{R1-Zero: Large Scale RL on Base Model}

Kimi K1.5 \cite{team2025kimi} and Kimi-VL \cite{team2025kimivl} employs a variant of the online policy mirror descent (OPMD) algorithm \cite{mei2019principled, tomar2020mirror} to apply RL to MLLMs, thereby enhancing their reasoning capabilities. Additionally, it adopts a length penalty mechanism to prevent the model from overthinking. This represents a successful attempt in the direction of MLLM reasoning based on RL.
Meanwhile, DeepSeek R1 \cite{guo2025deepseek} introduces a key innovation by validating that RL with Verifiable Reward (RLVR) is sufficient to effectively elicit reasoning capabilities in LLMs. The core of RLVR lies in its rule-driven reward mechanism, which optimizes model performance efficiently and reliably. Compared to traditional SFT and process supervision methods, RLVR demonstrates significant advantages when handling verifiable tasks, such as mathematical problem-solving and code generation: it eliminates the need for labor-intensive data generation and annotation while significantly improving model performance with limited training data. Additionally, by avoiding the complex training of reward models, RLVR effectively mitigates the issue of reward hacking.

Given an input question $q$, a policy model $\pi_\theta$, and its generated response $o$, RLVR evaluates the generated response using a verifiable reward function $R(q, o)$. The key role of this reward function is to determine whether the generated output is correct and assign a binary score:
\begin{align}
    R(q, o) = 
\begin{cases} 
1, & \text{if } o == \text{ground truth}, \\
0, & \text{otherwise}.
\end{cases}
\end{align}
In DeepSeek R1, the reward model is designed as follows:
\begin{align}
\label{eqa:r1}
    R = R_{acc} + \lambda R_{format},
\end{align}
where $R_{format}$ evaluates whether the output of the policy model $\pi_\theta$ adheres to predefined template specifications. In DeepSeek R1, the template may require the output to follow the format <think></think><answer></answer>, where <think></think> contains the model's reasoning process, and <answer></answer> includes the final result. On the other hand, $R_{acc}$ focuses on assessing whether the output in the <answer></answer> matches the ground truth. Through this design, RLVR not only ensures the accuracy of the model's output but also guides the model to generate structured and interpretable reasoning content, thereby enhancing both performance and explainability. 
In this survey, we refer to the method of using only the format reward and accuracy reward during the RL phase, as defined in Equation \ref{eqa:r1}, as the R1 training paradigm. 

\paragraph{R1: RL for Reasoning with Cold Start}

While RL has shown promise in improving the reasoning abilities of LLMs, relying solely on large-scale RL is insufficient for robust reasoning development. 
A fundamental challenge is that reinforcement learning algorithms often require a reasonably good initialization to function effectively. In reasoning tasks, the combinatorial nature of the action space makes it difficult to define intermediate rewards. Consequently, without a meaningful starting policy, exploration becomes inefficient, leading to excessively sparse rewards. This frequently causes training to collapse into trivial or degenerate behaviors.
Cold-start methods address this challenge by providing initial scaffolding, including techniques such as curriculum design, synthetic demonstrations, weak supervision, or structured exploration, to bootstrap the model into a regime where RL can be effectively applied. DeepSeek R1 collected thousands of cold-start data to fine-tune the base model, enhancing the readability of reasoning CoT and the potential for RL. Thus, successful reasoning training typically involves a hybrid approach: cold start strategies to seed useful reasoning behaviors, followed by RL to refine and optimize them over time.

\subsubsection{R1 Training Paradim for MLLM}
Inspired by the design of format reward and accuracy reward mechanisms in Equation \ref{eqa:r1} within DeepSeek R1, researchers began attempting to transfer this successful experience to multimodal tasks.

\paragraph{MLLM-R1-Zero}

MedVLM-R1 \cite{pan2025medvlm} extended the DeepSeek R1 training paradigm to visual question-answering (VQA) tasks in the medical domain, guiding the model to answer medical multiple-choice questions through explicit reasoning paths, thereby improving prediction accuracy and OOD generalization performance. Similarly, VisualThinker-R1-Zero \cite{zhou2025r1} and MM-Eureka \cite{meng2025mm} successfully replicated DeepSeek R1's long CoT characteristics, including the "Aha Moment," self-reflection mechanisms, and dynamic adjustment of response length, in spatial reasoning and mathematical VQA tasks, respectively. These features demonstrate that the experience of DeepSeek R1 can be effectively transferred to complex multimodal reasoning tasks. OThink-MR1 \cite{liu2025othink} improved the widely used GRPO algorithm by introducing dynamic weighting for the KL divergence term, balancing exploration in the early training stages with exploitation in later stages. To filter data with appropriate difficulty levels for RL training, ThinkLite-VL \cite{wang2025sota} proposes a sample selection method based on MCTS. This method quantifies the difficulty of each sample by the number of iterations required for the VLM to solve the problem.
Despite these efforts achieving the application of the R1 training paradigm in multimodal tasks, they remain constrained by Equation \ref{eqa:r1} format rewards and accuracy rewards, somewhat neglecting the unique characteristics of tasks in the multimodal domain.

\paragraph{MLLM-R1}

To improve the stability of the RL training process and the upper limit of the model's inference performance, some methods have begun constructing multi-modal SFT data to cold-start the base model. 
Vision-R1 \cite{huang2025vision}  first leverages VLM and DeepSeek R1 to construct reasoning cold-start data and performs SFT. Subsequently, it employs Progressive Thinking Suppression Training (PTST), a staged RL strategy that incrementally extends the length of CoT reasoning while mitigating overthinking from the early stages.
Due to the relative scarcity of image-text data compared to pure text data, LMM-R1 \cite{peng2025lmm} proposes a two-stage training strategy to make efficient use of textual CoT data: the first stage involves RL training on pure text data to enhance MLLM's foundational text reasoning abilities, while the second stage incorporates image-text data to further improve the model's generalization capabilities.
R1-Onevision \cite{yang2025r1} highlights that long reasoning paths may cause the model to overlook visual information. In the SFT stage, R1-Onevision constructs a reasoning dataset incorporating formalized image descriptions. By standardizing the model's reasoning paths, it enables the model to actively integrate multimodal information during the reasoning process. In the RL stage, the model's reasoning capabilities and generalization are further strengthened.
Skywork R1V \cite{skywork2025r1v} proposes a lightweight R1 paradigm training method to alleviate the high consumption of hardware resources. This approach combines a visual encoder with a reasoning-capable LLM backbone through a lightweight visual projection module. The module is iteratively trained using SFT and the R1 paradigm to enhance the reasoning performance of MLLMs. 
Unlike cold-starting with SFT, Skywork R1V2 \cite{wei2025skywork} employs MPO for model cold-starting to address the issue where SFT undermines the performance of subsequent reinforcement learning or reasoning processes. Additionally, a Selective Sample Buffer (SSB) mechanism is designed to address the "vanishing advantage" problem in GRPO.
To address the diversity of mathematical problems, Reason-RFT \cite{tan2025reason} designed customized reward mechanisms for various problems. Each mechanism is tailored to the unique characteristics of its respective problem category, ensuring precise and fair evaluations. 

\paragraph{MLLM-R1 for Other Modalities}
Beyond image-text interactions, some research has explored applying the R1 training paradigm to tasks involving other modalities. R1-Omni \cite{zhao2025r1} conducted preliminary explorations in mixed-modality tasks involving video and audio. It first performs SFT cold-start with a small amount of reasoning data, then utilized the R1 training paradigm to explore the application of reasoning models in emotion recognition tasks. Spatial-R1 \cite{ouyang2025spatial} extends the R1 training paradigm to the field of video spatial reasoning. Additionally, R1-AQA \cite{li2025reinforcement}, SARI \cite{wen2025sari} and R1-Zero-VSI \cite{liao2025improved} have conducted a series of reinforcement learning-related explorations in audio understanding and video spatial reasoning tasks, respectively. These efforts provide new insights for the further development of multimodal reasoning.

\subsection{Reward Mechanism Design for Multimodal Perception}
\label{subsec:rm4mllm}

\subsubsection{Outcome-supervised reward Mechanism}

Outcome-supervised reward mechanisms represent an  advancement in aligning RL with complex, multimodal tasks by leveraging task-specific and cross-modal feedback. These mechanisms not only enhance model performance through structured and hierarchical rewards but also pave the way for more interpretable and generalizable reasoning across diverse modalities. We summarize the reward functions of different reward strategies in Table \ref{tab:rewerd-task-oriented}, Table \ref{tab:rewerd-cross-modal}, and Table \ref{tab:rewerd-curriculum}, respectively.

\paragraph{Task-Oriented Reward Strategy}

\begin{table}[t!]
\centering
\resizebox{0.8\textwidth}{!}{
\begin{tabular}{l|c}
\toprule
Model                      & Reward\\
\midrule
Visual-RFT \cite{liu2025visual} & \begin{tabular}[c]{@{}c@{}}Detection Tasks: $R=r_{IoU}+r_{conf}+r_{format}$\\ Classification Task:$R=r_{acc}+r_{format}$\end{tabular}\\ \midrule
Seg-Zero \cite{liu2025seg} & $R=r_{BIoU} + r_{BL1} + r_{PL1} + r_{SFormat} + r_{format}$\\ \midrule
VLM-R1 \cite{shen2025vlm} & \makecell[c]{$R_{acc}^{rec}(q,o)=\mathrm{IoU}(b^*,f_{rec}(o))$, $s_{ovd} =\min(1,\frac{L_{gt}}{L_{pred}})$ \\ $R_{acc}^{ovd}(q,o) =s_{ovd}\cdot\mathbf{mAP}(\mathbf{b}_{pred},\mathbf{b}_{gt})$} \\ \midrule
Perception-R1 \cite{yu2025perception} & \makecell[c]{$r_{\mathrm{answer}}=\frac{1}{N}\sum_{i=1}^N\Phi\left(y_i,z_{\hat{\sigma}(i)}\right)$, $R_{\mathrm{total}}=r_{\mathrm{format}}+r_{\mathrm{answer}}$} \\ \midrule
R1-SGG \cite{chen2025compile} & \makecell[c]{$R_{\text{total}} = R_{\text{format}} + R_{\text{node}} + R_{\text{edge}}$, \\ $R_{\text{node}} = 
\lambda_1 \cdot \text{Sim}(c_i, \tilde{c}_j) 
+ \lambda_2 \cdot \text{IoU}(b_i, \tilde{b}_j)
+ \lambda_3 \cdot \exp(-\|b_i - \tilde{b}_j\|_1)$ \\ $R_{\text{edge}} = 
\text{Sim}(v_i, \tilde{v}_k) \cdot 
\text{Sim}(v_j, \tilde{v}_l) \cdot 
\text{Sim}(p_{ij}, \tilde{p}_{kl})$} \\ \midrule
VideoChat-R1 \cite{li2025videochat} & \makecell[c]{ 
$R_{\text{st}} = R_{\text{format}} + R_{\text{IoU}}$, 
$R_{\text{qa}} = R_{\text{format}} + R_{\text{accuracy}}$\\
$R_{\text{gqa}} = R_{\text{format}} + R_{\text{IoU}} + R_{\text{accuracy}}$,
$R_{\text{cap}} = R_{\text{format}} + R_{\text{recall}}$}
\\ \midrule
TinyLLaVA-Video-R1 \cite{zhang2025tinyllava} & \makecell[c]{
$FR = r_0 + LR$, $
LR = \min\left(1, \frac{\text{Len}}{\text{ML}} \right) \cdot r_1$ \\
$AR = \begin{cases} r_2, & \text{if }o\text{ is correct} \\ 0, & \text{otherwise} \end{cases}$ \\ $R =
\begin{cases}
AR + FR, & \text{if } FR > 0 \text{ and } AR = r_2 \\
-FR, & \text{if } FR > 0 \text{ and } AR = 0 \\
-(r_0 + r_1 + r_2), & \text{if } FR = 0
\end{cases}$
}
\\ \bottomrule
\end{tabular}}
\caption{Reward function of RL-Based MLLM reasoning methods for Task-Oriented Reward Strategies.}
\label{tab:rewerd-task-oriented}
\end{table}

Given the complexity and diversity of multimodal tasks, recent studies have explored reward designs that leverage the intrinsic properties of each modality to broaden the applicability of RL.
Visual-RFT \cite{liu2025visual} targets fundamental computer vision problems such as image classification, object detection, and image-text grounding, introducing task-specific visual rewards. All of these rewards are based on the final prediction accuracy, providing clear label supervision signals within the visual modality. 
Seg-Zero \cite{liu2025seg} extends this concept to the field of visual segmentation. In addition to pixel-level IoU and L1 distance metrics used for evaluating segmentation quality, this method introduces model-generated CoT responses as part of the output, enabling partial supervision of the reasoning trajectory.
Similarly, VLM-R1 \cite{shen2025vlm} designs reward functions for referring expression comprehension and open-vocabulary object detection. To address the multi-agent attributes in visual perception tasks, Perception-R1 \cite{yu2025perception} formulates the reward matching problem as a bipartite graph matching task and employs the Hungarian algorithm \cite{kuhn1955hungarian} to maximize the overall reward. This ensures that each predicted attribute is accurately matched to its corresponding ground truth, thereby optimizing the reward calculation process.
R1-SGG \cite{chen2025compile} and VideoChat-R1 \cite{li2025videochat} both employ modular reward functions, tailored for structured scene graph generation and spatiotemporal video tasks, respectively. They design composite rewards—covering format, node/edge alignment, IoU, and task-specific correctness—to guide model outputs toward structured, temporally grounded responses. In contrast, TinyLLaVA-Video-R1 \cite{zhang2025tinyllava} applies a lightweight reward scheme combining answer correctness, format adherence, and reasoning length, enabling small models to exhibit interpretable rationales under limited supervision. In addition to designing rewards for single-modality characteristics, some methods have begun to consider cross-modal interactions, representing a transitional phase in the broader shift from single-modality to multi-modality reward modeling.

\paragraph{Cross-Modal Interaction Reward Strategy}

\begin{table}[t!]
\centering
\resizebox{0.8\textwidth}{!}{
\begin{tabular}{l|c}
\toprule
Model                      & Reward\\
\midrule
Q-Insight \cite{li2025q} & $r^{(i)} = r^{(i)}_{\text{fmt}} + \mathbf{1}_{\text{scr}} \cdot r^{(i)}_{\text{scr}} + \mathbf{1}_{\text{deg}} \cdot \left( \alpha_1 \cdot r^{(i)}_{\text{deg}} + \alpha_2 \cdot r^{(i)}_{\text{lev}} \right)$ \\ \midrule
VLAA-Thinker \cite{chen2025sft} & \makecell[c]{$R=r_{acc} + \lambda r_{format}$ \\ $R_{open}=1-\exp(-(S_\theta(\hat{y})-S_\theta(y))\times\beta)\mathrm{~if~}f_\theta(\hat{y})>f_\theta(y)\mathrm{~else~}0$} \\  \midrule
Relation-R1 \cite{li2025relation} & \makecell[c]{$r_{\mathrm{form}}(o_i)=
\begin{cases}
1 & \mathrm{if}~o_i\text{ adheres to the format}, \\
0 & \mathrm{otherwise}.  
\end{cases}$ \\ $r_\mathrm{binary}{(o)}=\alpha\cdot Recall+(1-\alpha)\cdot mRecall$ \\ $r_{\mathrm{n-ary}}(o)=\beta\cdot Acc_v+(1-\beta)\cdot Acc_n$} \\ \midrule
FAST \cite{xiao2025fast} & \makecell[c]{$r_a=\mathbb{I}(A_{pred}\equiv A_{gt})=
\begin{cases}
1.0 & \mathrm{if~}A_{pred}\equiv A_{gt} \\
0.0 & \mathrm{otherwise}
\end{cases}$ \\ $r_f=\mathbb{I}_{\text{think-tags}}\cdot\mathbb{I}_{\text{answer-tags}}$ \\ $r_t=
\begin{cases}
1-\frac{L}{L_{avg}} & \inf S_{\mathrm{comp.}}<\theta_{comp.}\mathrm{~and~}r_a=1 \\
min(\frac{L}{L_{avg}}-1,1) & \inf\theta_{comp.}\leq S_{\mathrm{comp.}}\mathrm{~and~}r_a=0 \\
0 & \mathrm{otherwise} 
\end{cases}$} \\ \midrule
Video-R1 \cite{feng2025video} & \makecell[c]{ $r_t= \begin{cases} \alpha, & \mathrm{if~}p>\mu\cdot\tilde{p} \\ 0, & \mathrm{otherwise} \end{cases},$ $r_i^{\mathrm{T-GRPO}}= \begin{cases} r_i+r_t, & \mathrm{if~}o_i\mathrm{~is~correct} \\ r_i, & \mathrm{otherwise} \end{cases}$ \\ $r_i= \begin{cases} r_i+\omega, & \mathrm{if~}o_i\text{ is correct and }l_{\min}\leq\mathrm{len}(o_i)\leq l_{\max} \\ r_i, & \mathrm{otherwise} \end{cases}$ } \\ \midrule
MetaSpatial \cite{pan2025metaspatial} & \makecell[c]{ $R_\mathrm{physics}=-\alpha\cdot\text{CollisionRatio}-\beta\cdot\text{ConstraintRatio}$\\  $R_\mathrm{render}=\frac{1}{50}\sum_{i=1}^5\mathrm{Grade}_i, $  $R=\lambda_1r_\mathrm{format}+\lambda_2r_\mathrm{physics}+\lambda_3r_\mathrm{render}$ }\\ \midrule
GFlowVLM \cite{kang2025gflowvlm} & \makecell[c]{NumberLine:$R(x) = R(c, y_t) = \frac{l}{|c - y_t| + 1}, \quad \text{where } l = 100$\\ BlackJack: $R(x) = \max\left(1 \times 10^{-10}, \; (r(x) + 1) \times 10\right)$\\ ALFWorld: $R(s_t, a_t, s_{t+1} \mid g_{\text{task}}) = 50 \cdot \mathbf{1}\{s_{t+1} = g_{\text{task}}\} + \mathbf{1}\{s_{t+1} \in \text{subgoals}\}$ }
\\ \bottomrule
\end{tabular}}
\caption{Reward function of RL-Based MLLM reasoning methods for Cross-Modal Interaction Reward Strategies.}
\label{tab:rewerd-cross-modal}
\end{table}

To address the non-linear interactions in multimodal reasoning, recent studies assign modality-specific rewards to promote more active cross-modal interactions.
Q-Insight \cite{li2025q} employs GRPO to jointly optimize score accuracy and distortion classification, enabling perceptual reasoning with minimal supervision and strong zero-shot generalization. Building on this, VLAA-Thinker \cite{chen2025sft} introduces a mixed reward module combining rule-based and open-ended signals to guide adaptive, realistic reasoning. Relation-R1 \cite{li2025relation} targets complex visual relations by designing Binary and N-ary Relation Rewards, emphasizing visual-semantic grounding over linguistic bias. To counter overthinking, Fast \cite{xiao2025fast} introduces a complexity-aware thinking reward and adaptively adjusts the KL coefficient based on visual and semantic difficulty. Some methods extend RL to temporal reasoning via audio–visual modeling. Video-R1 \cite{feng2025video} introduces a reward based on temporal consistency by comparing predicted and ground-truth frame orders, guiding causal inference across time. While the reward remains output-based and coarse-grained, it marks a step toward RL-driven multimodal temporal understanding.

As reinforcement learning extends to spatial reasoning, reward design must accommodate not only semantic alignment across modalities but also reasoning over actions, layouts, and 3D constraints. MetaSpatial \cite{pan2025metaspatial} introduces a multi-tier reward system for 3D scene generation, combining format validation, physics-based verification, and GPT-4o-driven perceptual scoring. This hierarchical structure enables dense, cross-modal feedback and supports grounded scene generation. GFlowVLM \cite{kang2025gflowvlm} further advances process-level modeling using GFlowNets to sample reasoning trajectories with probabilities aligned to structured rewards. It employs Trajectory-, Subtrajectory-, and Detailed-Balanced losses to optimize coherence at both local and global levels, enabling fine-grained reward propagation throughout multimodal inference.

\paragraph{Curriculum-Based Reward Strategy}

\begin{table}[t!]
\centering
\resizebox{0.8\textwidth}{!}{
\begin{tabular}{l|c}
\toprule
Model                      & Reward\\
\midrule
Curr-ReFT \cite{deng2025boosting} & \makecell[c]{
$\mathbf{R_{Binary}}(\mathbf{o}_{std},\mathbf{o}_{gt})= \begin{cases}
1, & \mathrm{if~}\mathbf{o}_{std}=\mathbf{o}_{gt} \\
0, & \mathrm{otherwise} 
\end{cases},$ 
$\mathbf{R}_{s}(\mathbf{o}_{std},\mathbf{o}_{gt})=
\begin{cases}
1, & \mathbf{o}_{std}=\mathbf{o}_{gt} \\
0, & \mathrm{otherwise}
\end{cases}$ \\
$\mathbf{R}_{m}(\mathbf{o}_{std},\mathbf{o}_{gt})=
\begin{cases}
1, & \mathbf{o}_{std}=\mathbf{o}_{gt} \\
0.2, & \mathbf{o}_{std}\subset\mathbf{o}_{gt},|\mathbf{o}_{std}|>0 \\
0, & \mathrm{otherwise}
\end{cases}, $ \\
$\mathbf{R}_{acc\_cls}=\frac{|P\cap G|}{|P\cup G|}=\frac{|\{c_{i}|c_{i}\in P\mathrm{~and~}c_{i}\in G\}|}{|\{c_{1},...,c_{m}\}\cup\{g_{1},...,g_{n}\}|}, $ $\mathbf{R}_{det}=\mathbf{R}_{acc\_det}+\mathbf{R}_{format}$\\
} \\ \midrule
Embodied-R \cite{zhao2025embodied} & \makecell{$r_i = \omega_1 \cdot r^{\text{format}}_i + \omega_2 \cdot r^{\text{accuracy}}_i + \omega_3 \cdot r^{\text{logic}}_i$} \\ \midrule
NoisyRollout \cite{liu2025noisyrollout} & \makecell{
$R_i=
\begin{cases}
1, & \mathrm{if~}o_i\mathrm{~is~correct} \\
0, & \mathrm{otherwise}
\end{cases}$, $\bar{R}=\frac{1}{n_1+n_2}\sum_{i=1}^{n_1+n_2}R_i$
}
\\ \bottomrule
\end{tabular}}
\caption{Reward function of RL-Based MLLM reasoning methods for Curriculum-Based Reward Strategy.}
\label{tab:rewerd-curriculum}
\end{table}

In addition to adopting the standard R1 training paradigm, Curr-ReFT \cite{deng2025boosting} employs a three-stage curriculum consisting of binary classification, multiple-choice selection, and open-ended question answering, with each stage guided by task-specific rewards to progressively build reasoning ability. Embodied-R \cite{zhao2025embodied} also adopts a staged training approach, introducing different weights for $\omega$ across training phases. Additionally, it incorporates a logical consistency reward to align the reasoning process with the final answer in spatial tasks. NoisyRollout \cite{liu2025noisyrollout} enhances exploration by mixing clean and noisy visual inputs, using a noise annealing schedule to guide training from early diversity to stable convergence.

\subsubsection{Process-Supervised Reward Mechanism}
To mitigate sparse feedback and improve structural reasoning, recent studies adopt process-based rewards to enhance logical coherence.
R1-VL \cite{zhang2025r1} serves as a representative example of this direction. It introduces Step-wise Group Relative Policy Optimization (StepGRPO), an online reinforcement learning framework that incorporates two structurally grounded reward components designed to operate over the reasoning process: StepRAR evaluates whether the model’s output includes key intermediate reasoning steps, while StepRVR assesses the logical coherence and completeness of the overall reasoning chain. These rewards are anchored at intermediate states and span the entire vision–language–reasoning trajectory, significantly improving the model’s ability to maintain logical consistency in complex tasks.  Similarly, ChestX-Reasoner \cite{fan2025chestx} introduces a process reward based on clinical reports, evaluating the factual accuracy of intermediate reasoning steps to enhance diagnostic coherence and reasoning quality.

\subsection{Training Efficiency and Stability}
\label{subsec:effAndStab}

Training efficiency and stability are critical in the development of robust reasoning models, particularly in complex multimodal settings. By leveraging strategies such as curriculum learning, sample efficiency optimization, and techniques to mitigate catastrophic forgetting, researchers can enhance both the speed and reliability of model training while preserving previously acquired knowledge.

\subsubsection{Curriculum Learning}
Curriculum learning is a training strategy that emulates the human learning process by organizing tasks in an order of increasing difficulty, allowing models to first acquire basic skills before advancing to more complex reasoning abilities. In RL, especially for multimodal reasoning tasks, curriculum learning helps mitigate training instability and accelerates convergence by providing structured, progressively challenging learning stages.
Kimi K1.5 \cite{team2025kimi} adopts curriculum and prioritized sampling to expose the model to increasingly difficult examples while focusing on weak areas, enhancing overall training efficiency. Curr-ReFT \cite{deng2025boosting} introduces difficulty-aware reward shaping to stage the learning process, facilitating stable reward-based training.
Beyond data and reward curricula, Embodied-R \cite{zhao2025embodied} implements a three-stage RL schedule that gradually shifts reward weights from format to accuracy and logical consistency, guiding the model toward coherent embodied reasoning. NoisyRollout \cite{liu2025noisyrollout} implicitly applies curriculum via a noise annealing schedule—starting with distorted visual inputs and gradually reducing noise—promoting early exploration and stable convergence without added cost.

\subsubsection{Sample Efficiency}

RL-based reasoning heavily relies on high-quality samples to guide effective policy optimization. Insufficient or unreliable samples can lead to slow convergence and suboptimal performance, ultimately impairing reasoning and generalization.
Skywork R1V \cite{skywork2025r1v}  improves sample efficiency via Adaptive-Length CoT Distillation, which adjusts reasoning chain length to reduce computation, and a Hybrid Optimization Framework that iteratively refines the model using both high-confidence and error samples, reducing dependence on large-scale annotations.
MM-Eureka \cite{meng2025mm} proposes a two-stage training strategy, with the second stage focusing on enhancing the model using domain-specific data. Additionally, the authors introduce the MMK12 \footnote{https://huggingface.co/datasets/FanqingM/MMK12} multimodal mathematical reasoning dataset to highlight the importance of high-quality data. 
GFlowVLM \cite{kang2025gflowvlm} adopts generative flow networks to sample and learn from high-value reasoning trajectories, enabling efficient off-policy training and promoting diverse yet meaningful exploration. 
LMM-R1 \cite{peng2025lmm} employs a two-stage method, FRE followed by MGT, where the model is first trained on textual data to enhance reasoning and then generalized to other modalities, thereby mitigating the scarcity of high-quality multimodal data.
MetaSpatial \cite{pan2025metaspatial} introduces a GRPO-based multi-turn refinement strategy, selecting higher-quality grouped trajectories to accelerate spatial reasoning.
NoisyRollout \cite{liu2025noisyrollout} demonstrates strong generalization with only 2.1K samples by mixing clean and noisy rollouts, enhancing learning through visual diversity at no extra cost.

\subsubsection{Catastrophic forgetting}

Catastrophic forgetting refers to the phenomenon where models, after fine-tuning on new tasks, lose previously acquired abilities.
SFT often induces catastrophic forgetting by overwriting prior capabilities, whereas RL, by optimizing reward signals rather than explicit outputs, better preserves general reasoning skills \cite{liu2025seg, peng2025lmm}.  
In addition to the inherent advantages of RL, the mainstream solution in current RL-based methods to mitigate this issue is to incorporate a KL regularization term, which helps prevent excessive parameter drift \cite{pan2025medvlm, liu2025othink, team2025kimi}. Curr-ReFT \cite{deng2025boosting} proposes Rejected Sample-based Self-improvement, a method that selectively learns from high-quality multimodal and textual examples to maintain the fundamental capabilities of MLLMs. This approach ensures that while enhancing the model's ability to perform new tasks, it retains its original knowledge and skills.

\subsection{RLHF Training Framework}
\label{subsec:RLHF-Framwork}

\begin{table}[t!]
\begin{tabular}{l cc}
\toprule
Framework      & Models  & Algorithm \\ \midrule
DeepSpeed-Chat \cite{rasley2020deepspeed} & LLM     & PPO \\
OpenRLHF \cite{hu2024openrlhf} & LLM     & GRPO, PPO, REINFORCE++, RLOO, etc \\
LLaMA-Factory \cite{zheng2024llamafactory}  & LLM     & PPO, DPO, KTO, ORPO, etc \\
TRL \cite{vonwerra2022trl} & MLLM    & GRPO, PPO, ORPO, RLOO, etc \\
ColossalAI \cite{ColossalAI} & MLLM    & GRPO, PPO, DPO, KTO, etc \\
VeRL \cite{sheng2024hybridflow} & MLLM    & GRPO, DAPO, PPO, RLOO, etc \\
Open-R1  \cite{openr1} & LLM     & GRPO \\
R1-V  \cite{chen2025r1v} & MLLM    & GRPO \\
EasyR1 \cite{zheng2025easyr1} & MLLM & GRPO, REINFORCE++, RLOO, ReMax \\ \bottomrule 
\end{tabular}
\caption{Overview of RLHF frameworks with supported models and RL algorithms.}
\label{tab:framework}
\end{table}

In recent years, the rise of RLHF and reward learning has led to the emergence of various open-source libraries that lower research barriers and enhance development efficiency. Table \ref{tab:framework} summarizes major RLHF frameworks and their supported algorithms. DeepSpeed-Chat \cite{rasley2020deepspeed} enables efficient, cost-effective ChatGPT-style model training. OpenRLHF \cite{hu2024openrlhf} supports distributed training for large models and integrates with HuggingFace. LLaMA-Factory \cite{zheng2024llamafactory} offers broad model and training method support with a visual interface. TRL \cite{vonwerra2022trl} focuses on post-training techniques like PPO and DPO with distributed support. Colossal-AI \cite{ColossalAI} simplifies large-scale training with parallel strategies. VeRL \cite{sheng2024hybridflow} targets multimodal and vision-language tasks with modular design. Open-R1 \cite{openr1} and its fork EasyR1 \cite{zheng2025easyr1} replicate and extend the DeepSeek-R1 pipeline. R1-V \cite{chen2025r1v} emphasizes long-chain reasoning in vision-language models. Together, these frameworks offer diverse capabilities that accelerate innovation in RLHF research.

\section{Applications with RL-Based MLLM Reasoning}
\label{sec:application}
RL has substantially expanded the application boundaries of MLLMs across diverse domains. By enhancing the structured reasoning ability of MLLMs through optimized reward mechanisms and policy improvements, RL-based methods enable models to adapt to increasingly complex real-world tasks. In this section, we categorize the emerging applications of RL-based MLLM reasoning into three major areas: Embodied AI, MLLM Agentic Systems, and Domain-Specific Applications of RL-Based Multimodal Reasoning.

\subsection{Embodied AI}
Embodied AI focuses on enabling MLLMs to perceive, reason, and act in physical or simulated environments. RL-based multimodal reasoning methods have become pivotal in improving models’ ability to interpret multimodal inputs and to generate action-oriented reasoning strategies.  Models such as MetaSpatial \cite{pan2025metaspatial} demonstrate the capability of integrating physics-based rewards and perceptual feedback to support three-dimensional scene understanding and object manipulation. Similarly, Video-R1 \cite{feng2025video} extends reasoning into the temporal domain, enabling video-based agents to infer causal relationships between sequential frames, a critical aspect for embodied perception in dynamic environments. Embodied-R \cite{zhao2025embodied} proposes a collaborative RL framework that activates embodied spatial reasoning by decoupling perception and reasoning, achieving superior performance and generalization with a logical consistency reward. The reinforcement learning framework facilitates iterative learning of spatial, temporal, and causal relations, allowing embodied MLLMs to achieve stronger generalization and physical realism. Autonomous driving can also be viewed as a representative embodied scenario, where RL-enhanced MLLMs hold promise for future advances in spatial-temporal reasoning and decision-making.

\subsection{MLLM Agentic System}
As MLLMs evolve beyond passive perception and generation, a critical frontier lies in the development of agentic systems—models capable of goal-directed reasoning, autonomous decision-making, and interaction with dynamic environments.  An agentic MLLM is expected not only to understand multimodal contexts but also to proactively plan, act, and adapt based on evolving task demands. RL provides a natural framework for cultivating such capabilities, by treating reasoning and interaction as sequential decision processes.

In interactive scenarios such as GUI-based task execution, UI-R1 \cite{lu2025ui} demonstrates the importance of multimodal action reasoning. By jointly optimizing rewards for action type prediction, argument selection, and output formatting, the model effectively learns to perform sequences of operations, aligning its outputs with human-intended goals. GUI-R1 \cite{xia2025gui} enhances the GUI capabilities of MLLM in advanced real-world task scenarios by unifying action types, input text, and click coordinates into a standardized action space framework. InfiGUI-R1 \cite{liu2025infigui} advances GUI agents from reactive execution to deliberative reasoning via a two-stage RL framework, enhancing planning and error recovery through sub-goal guidance and reflective correction.

\subsection{Domain-Specific Applications of RL-Based Multimodal Reasoning}
Beyond general embodied and agentic capabilities, RL–based multimodal reasoning has been increasingly applied to a range of specialized real-world domains that demand high levels of perception and decision-making under complex conditions. By leveraging structured rewards and sequential learning frameworks, these methods enable MLLMs to move beyond static understanding toward dynamic, context-sensitive behavior. Domains such as medical and healthcare, as well as human-centered interaction, serve as critical testbeds for evaluating and advancing the robustness, generalization, and interpretability of RL-empowered multimodal systems.

\subsubsection{Medical and Healthcare}
The medical and healthcare domains present unique demands for high-stakes reasoning, interpretability, and generalization. RL-based multimodal reasoning methods have shown promising progress in tasks such as medical visual question answering and clinical decision support. MedVLM-R1 \cite{pan2025medvlm} improves answer verifiability by rewarding structured reasoning traces for multiple-choice tasks. ChestX-Reasoner \cite{fan2025chestx} applies process-supervised reinforcement learning to align model reasoning with clinical workflows, using supervision signals extracted from radiology reports. This approach enhances factuality, completeness, and diagnostic relevance of generated reasoning chains, while also improving outcome accuracy across diverse tasks such as disease classification, anomaly detection, and temporal comparison.

\subsubsection{Social and Human}
Understanding human behavior, emotions, and social interactions is an emerging frontier for RL-based multimodal reasoning. R1-Omni \cite{zhao2025r1} integrates audio, video, and text to enhance emotional recognition via RL, enabling structured socio-emotional reasoning. Similarly, R1-AQA \cite{li2025reinforcement} trains MLLMs to interpret acoustic signals for auditory reasoning. These capabilities support empathetic AI agents and socially adaptive systems, where RL facilitates robustness, error recovery, and personalized interaction.

\section{Datasets and Benchmark}
\label{sec:dataAndBench}

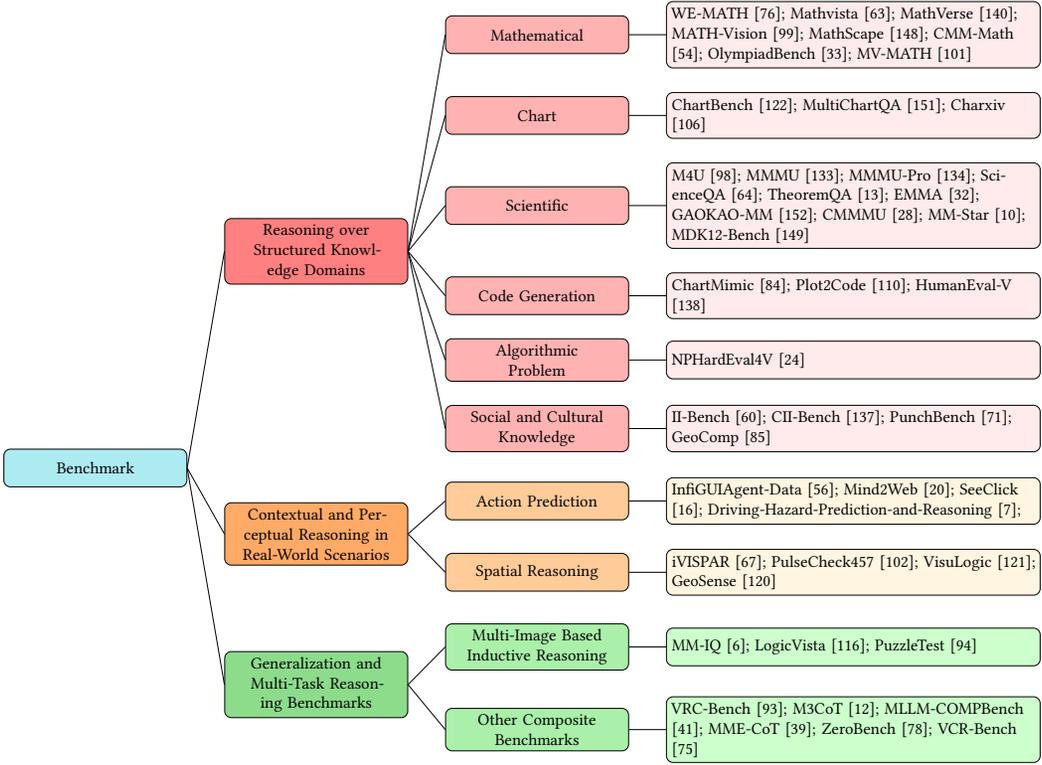
\begin{figure}[t!]
\centering
\resizebox{\textwidth}{!}{
\begin{tikzpicture}[
    node distance=0.5cm and 0.8cm,
    every node/.style={font=\normalsize, rounded corners},
    redbox/.style={draw=black, minimum height=0.8cm, text width=3.7cm, align=center},
    orangebox/.style={draw=black, minimum height=0.8cm, text width=3.7cm, align=center},
    greenbox/.style={draw=black, minimum height=0.8cm, text width=3.7cm, align=center},
    data/.style={fill=textgray, draw=black, minimum height=0.8cm, text width=7.8cm, align=left},
    root/.style={fill=mainblue, draw=black, minimum height=0.8cm, text width=3.7cm, align=center}
]

\node[root] (root) at (0,0) {Benchmark};

\node[redbox, fill=red3, right=of root, yshift=4.6cm] (domain) {Reasoning over Structured Knowledge Domains};
\node[orangebox, fill=orange3, right=of root, yshift=-1.4cm] (reasoning) {Contextual and Perceptual Reasoning in Real-World Scenarios};
\node[greenbox, fill=green3, right=of root, yshift=-4.6cm] (composite) {Generalization and Multi-Task Reasoning Benchmarks};

\node[redbox, fill=red2, right=of domain, yshift=4.6cm] (math) {Mathematical};
\node[redbox, fill=red2, below=of math, yshift=-0.4cm] (chart) {Chart};
\node[redbox, fill=red2, below=of chart, yshift=-0.6cm] (scientific) {Scientific};
\node[redbox, fill=red2, below=of scientific, yshift=-0.6cm] (code) {Code Generation};
\node[redbox, fill=red2, below=of code] (algo) {Algorithmic\\Problem};
\node[redbox, fill=red2, below=of algo] (social) {Social and Cultural\\Knowledge};

\node[data, fill=red1, right=of math] (mathd) {WE-MATH \cite{qiao2024we}; Mathvista \cite{lu2023mathvista}; MathVerse \cite{zhang2024mathverse}; MATH-Vision \cite{wang2024measuring}; MathScape \cite{zhou2024mathscape}; CMM-Math \cite{liu2024cmm}; OlympiadBench \cite{he2024olympiadbench}; MV-MATH \cite{wang2025mv}};
\node[data, fill=red1, right=of chart] (chartd) {ChartBench \cite{xu2023chartbench}; MultiChartQA \cite{zhu2024multichartqa}; Charxiv \cite{wang2024charxiv}};
\node[data, fill=red1, right=of scientific] (scid) {M4U \cite{wang2024m4u}; MMMU \cite{yue2024mmmu}; MMMU-Pro \cite{yue2024mmmupro}; ScienceQA \cite{lu2022learn}; TheoremQA \cite{chen2023theoremqa}; EMMA \cite{hao2025can}; GAOKAO-MM \cite{zong2024gaokao}; CMMMU \cite{ge2024cmmmu}; MM-Star \cite{chen2024we}; MDK12-Bench \cite{zhou2025mdk12}};
\node[data, fill=red1, right=of code] (coded) {ChartMimic \cite{shi2024chartmimic}; Plot2Code \cite{wu2024plot2code}; HumanEval-V \cite{zhang2024humaneval}};
\node[data, fill=red1, right=of algo] (algod) {NPHardEval4V \cite{fan2024nphardeval4v}};
\node[data, fill=red1, right=of social] (sociod) {II-Bench \cite{liu2024ii}; CII-Bench \cite{zhang2024can}; PunchBench \cite{ouyang2024punchbench}; GeoComp \cite{song2025geolocation}};

\node[orangebox, fill=orange2, right=of reasoning, yshift=0.7cm] (action) {Action Prediction};
\node[orangebox, fill=orange2, below=of action, yshift=-0.2cm] (spatial) {Spatial Reasoning};
\node[data, fill=orange1, right=of action] (actiond) {InfiGUIAgent-Data \cite{liu2025infiguiagent}; Mind2Web \cite{deng2023mind2web}; SeeClick \cite{cheng2024seeclick}; Driving-Hazard-Prediction-and-Reasoning \cite{charoenpitaks2024exploring};};
\node[data, fill=orange1, right=of spatial] (spatiald) {iVISPAR \cite{mayer2025ivispar}; PulseCheck457 \cite{wang2025pulsecheck457}; VisuLogic \cite{xu2025visulogic}; GeoSense \cite{xu2025geosense}};

\node[greenbox, fill=green2, right=of composite, yshift=0.8cm] (multiimg) {Multi-Image Based\\Inductive Reasoning};
\node[greenbox, fill=green2, below=of multiimg, yshift=-0.3cm] (other) {Other Composite\\Benchmarks};
\node[data, fill=green1, right=of multiimg] (mid) {MM-IQ \cite{cai2025mm}; LogicVista \cite{xiao2024logicvista}; PuzzleTest \cite{toh2025jumping}};
\node[data, fill=green1, right=of other] (otherd) {VRC-Bench \cite{thawakar2025llamav}; M3CoT \cite{chen2024m}; MLLM-COMPBench \cite{kil2024mllm}; MME-CoT \cite{jiang2025mme}; ZeroBench \cite{roberts2502zerobench}; VCR-Bench \cite{qi2025vcr}};

\draw[-] (root.east) -- (domain.west);
\draw[-] (root.east) -- (reasoning.west);
\draw[-] (root.east) -- (composite.west);

\foreach \x in {math, chart, scientific, code, algo, social}
  \draw[-] (domain.east) -- (\x.west);

\foreach \x in {action, spatial}
  \draw[-] (reasoning.east) -- (\x.west);

\foreach \x in {multiimg, other}
  \draw[-] (composite.east) -- (\x.west);

\foreach \x/\y in {
  math/mathd,
  chart/chartd,
  scientific/scid,
  code/coded,
  algo/algod,
  social/sociod,
  action/actiond,
  spatial/spatiald,
  multiimg/mid,
  other/otherd}
  \draw[-] (\x.east) -- (\y.west);

\end{tikzpicture}
}
\caption{Multimodal Reasoning Benchmarks.}
\label{fig:benchmark}
\end{figure}

To comprehensively evaluate the reasoning capabilities of MLLMs, researchers have developed a variety of benchmark datasets spanning different reasoning paradigms and modalities. In this section, we categorize these datasets into three high-level domains: Reasoning over Structured Knowledge Domains, Contextual and Perceptual Reasoning in Real-World Scenarios, and Generalization and Multi-Task Reasoning Benchmarks, with each domain encompassing several sub-domains that reflect distinct types of knowledge, reasoning structures, or contextual challenges. An overview of these representative benchmarks is illustrated in Figure~\ref{fig:benchmark}. We summarize the performance of current open-source and closed-source reasoning models on mainstream reasoning benchmarks in Table \ref{tab:benchmark_data}. 
Among these, Mulberry \cite{yao2024mulberry}, LLaVA-CoT \cite{xu2024llava}, Insight-V \cite{dong2024insight}, and LLaVA-Reasoner \cite{zhang2024improve} are mainstream reasoning models that do not rely on RL, while the remaining models are RL-based.
From Table \ref{tab:benchmark_data}, it is evident that a significant portion of current evaluations is concentrated in the first domain, particularly in mathematical, scientific, and chart-based reasoning tasks. This focus reveals a fundamental limitation: the dominance of well-structured, mathematically grounded benchmarks leaves open the question of whether these models can generalize to more complex, open-ended, and perceptually rich scenarios. Tasks that require interaction with dynamic environments, spatial abstraction, multimodal alignment, or socio-cultural reasoning are underrepresented, despite being critical for real-world applications. To address this limitation, we further include a range of benchmarks under the latter two categories, covering diverse scenarios such as user interface action prediction, 6D spatial reasoning, visual-to-code generation, and culturally grounded multimodal understanding. These datasets provide a more comprehensive lens for evaluating MLLMs' reasoning abilities beyond traditional symbolic domains. In the following, we present a detailed analysis of representative benchmarks within each sub-domain.

\begin{table}[t!]
\resizebox{\textwidth}{!}{
\begin{tabular}{l|cccccccccc}
\toprule
Model                   & MathVista & MathVision & MathVerse & MMMU-Val & MM-Star & MME     & ChartQA & DynaMath & HallBench & MMVet \\ \midrule
OpenAI o1               & 71.0     & -         & -        & 77.3    & -       & -       & -       & -        & -         & -     \\
GPT-4o 0513             & 63.8     & 30.4      & 39.4     & 69.1    & 63.9    & 2329.0  & 85.7    & 63.7     & 55.0      & -     \\
Claude-3.5-snooet-1022  & 65.3     & 35.6      & -        & 66.4    & 62.2    & 1920.0  & 90.8    & 64.8     & 55.0      & -     \\ \midrule
QVQ-72B-Preview         & 71.4     & 35.9      & -        & 70.3    & -       & -       & -       & -        & -         & -     \\
Mulberry-7B             & 63.1      & -        & -        & 55.0    & 61.3    & 2396.0  & 83.9    & 45.1     & 54.1      & -     \\
LLaVA-CoT-11B           & 54.8      & -        & -        & -       & 57.6    & -       & -       & -        & -         & -     \\
Insight-V               & 59.9      & -        & -        & 50.2    & 61.5    & 1685.1  & 81.50   & -        & -         & -     \\
LLaVA-Reasoner          & 50.6      & -        & -        & -       & -       & -       & 83.00   & -        & -         & -     \\
Kimi K1.5-Long CoT      & 74.9     & 38.6      & -        & 70.0    & -       & -       & -       & -        & -         & -     \\
Kimi K1.5-Short CoT     & 70.1     & 31.0      & -        & 68.0    & -       & -       & -       & -        & -         & -     \\
Kimi-VL-A3B             & 68.7      & 21.4     & -        & 57.0    & 61.3    & -       & -       & -        & -         & 66.7  \\
Vision-R1-7B            & 73.5     & -         & 52.4     & -       & -       & -       & -       & -        & -         & -     \\
Vision-R1-LlamaV-CI-11B & 62.7     & -         & 27.1     & -       & 61.4    & 2190.0  & 83.9    & -        & 49.5      & -     \\
MM-Eureka-8B            & 67.1     & 22.2      & 40.4     & -       & -       & -       & -       & -        & -         & -     \\
Curr-ReFT-3B            & 58.6     & -         & -        & -       & -       & -       & -       & -        & -         & 29.9 \\
Curr-ReFT-7B            & 64.5     & -         & -        & -       & -       & -       & -       & -        & -         & 35.6 \\
LMM-R1: MGT-Multi       & 59.0     & 26.8      & 41.8     & -       & 54.4    & -       & -       & -        & -         & -     \\
LMM-R1: MGT-PerceReson  & 63.2     & 26.4      & 41.6     & -       & 58.03   & -       & -       & -        & -         & -     \\
R1-Onevision-7B         & 64.1     & 29.9      & 40.0     & -       & -       & -       & -       & -        & -         & -     \\
R1-VL-2B                & 52.1     & 17.1      & 26.2     & -       & 49.8    & 2048.0 & 75.2     & 29.4     & 44.0      & -     \\
R1-VL-7B                & 63.5     & 24.7      & 40.0     & -       & 60.0    & 2376.0 & 83.9     & 45.2     & 54.7      & -     \\
Skywork R1V             & 67.5     & -         & -        & 69.0    & -       & -       & -       & -        & -         & -     \\
ThinkLite-VL-7B         & 75.1     & 32.9      & 50.7     & 54.6    & 65.00   & -       & -       & -        & -         & 67.8  \\
VLAA-Thinker-Qwen2.5-3B & 61.0     & 24.4      & 36.4     & -       & -       & -       & -       & -        & -         & -     \\
VLAA-Thinker-Qwen2.5-7B & 68.0     & 26.4      & 48.2     & -       & -       & -       & -       & -        & -         & -     \\
VL-Rethinker-7B         & 74.9     & 32.3      & 54.2     & 56.7    & -       & -       & -       & -        & -         & -     \\
VL-Rethinker-72B        & 80.3     & 43.9      & 61.7     & 68.8    & -       & -       & -       & -        & -         & -     \\
Skywork R1V2            & 74.0     & 49.0      & -        & 73.6    & -       & -       & -       & -        & -         & -     \\
FAST-3B                 & 66.2     & 26.8      & 43.0     & -       & -       & -       & -       & 54.4     & -         & 64.0  \\
FAST-7B                 & 73.8     & 30.6      & 50.6     & -       & -       & -       & -       & 58.3     & -         & 71.2  \\
NoisyRollout            & 69.6     & 28.5      & 53.2     & -       & -       & -       & -       & -        & -         & -     \\ \bottomrule
\end{tabular}}
\caption{Performance of RL-Based MLLMs on Reasoning Benchmarks.}
\label{tab:benchmark_data}
\end{table}

\subsection{Reasoning over Structured Knowledge Domains}

This category encompasses benchmarks that evaluate reasoning within structured and well-defined knowledge domains, such as mathematics, science, and programming.  These tasks typically involve symbolic inputs and domain-specific visuals, including equations, diagrams, or structured textual prompts. The benchmarks are designed to assess whether MLLMs can perform formal symbolic manipulation, conduct multi-step logical inference, and abstract visual-textual relationships within highly regular and semantically rich contexts. 

\paragraph{Mathematical Reasoning.} A broad range of benchmarks have been proposed to assess the mathematical reasoning capabilities of multimodal large language models, with tasks that require precise alignment between symbolic understanding, visual interpretation, and logical deduction. MathVista \cite{lu2023mathvista}, MathVerse \cite{zhang2024mathverse}, and WE-MATH \cite{qiao2024we} focus on school-level mathematical problems presented in multimodal formats, including equation solving, geometric diagrams, and accompanying textual prompts. These datasets test models’ abilities to interpret structured mathematical content, navigate spatial relationships, and perform multi-step numeric reasoning. MATH-Vision \cite{wang2024measuring} and MV-MATH \cite{wang2025mv} extend this paradigm by transforming LaTeX-based mathematical expressions into images, evaluating whether models can comprehend symbolic structures purely from visual representations. OlympiadBench \cite{he2024olympiadbench} introduces high-difficulty problems drawn from math olympiads, emphasizing abstract reasoning and competition-level problem-solving that often requires multi-hop inference. 
The uniqueness of MathScape \cite{zhou2024mathscape} lies in its vision-centered, open-ended question design, which integrates abstract symbolic reasoning with contextual visual cues. This setting encourages models to transcend direct computation and fosters the development of analogy-based and relational understanding capabilities.
CMM-Math \cite{liu2024cmm} further incorporates CoT prompting into multimodal mathematical reasoning, providing structured intermediate steps to assess whether models can generate interpretable reasoning paths across both vision and language modalities.

\paragraph{Chart Reasoning.}
Chart-based reasoning benchmarks evaluate the ability of MLLMs to extract structured information from visual plots and perform quantitative analysis grounded in natural language. ChartBench \cite{xu2023chartbench} and MultiChartQA \cite{zhu2024multichartqa} present a wide range of questions over diverse chart types, such as bar graphs, pie charts, and line plots, requiring aggregation, comparison, or prediction over visual data. CharXiv \cite{wang2024charxiv} adopts a more challenging evaluation setting by sourcing all charts from real-world scientific publications. The benchmark incorporates complex domain-specific notations, multi-subplot layouts, and visually dense figures, as well as semantically ambiguous or unanswerable questions, thereby substantially increasing the task complexity. 

\paragraph{Scientific Reasoning.}
Scientific reasoning benchmarks aim to assess a model’s ability to integrate domain-specific visual information with conceptual understanding across disciplines such as physics, chemistry, and biology. MMMU \cite{yue2024mmmu} and MMMU-Pro \cite{yue2024mmmupro} provide large-scale multimodal assessments based on academic exam questions, with inputs that include circuit diagrams, biological illustrations, and chemical reaction graphs. These benchmarks demand subject-specific knowledge and cross-modal multi-hop inference. 
M4U \cite{wang2024m4u} extends this direction by constructing a a multilingual benchmark that covers 64 disciplines. Its tasks feature rich visual materials such as graphs, experimental setups, and symbolic content, paired with multilingual questions designed to test the deep scientific understanding, multilingual alignment, and multimodal reasoning capabilities of MLLMs in real-world educational settings. 
MDK12-Bench \cite{zhou2025mdk12} complements these efforts by focusing on structured academic reasoning across core school subjects, while incorporating dynamic evaluation to enhance robustness. ScienceQA \cite{lu2022learn} and TheoremQA \cite{chen2023theoremqa} offer grade-school to high-school level science problems that emphasize explanation and factual integration, often including illustrated contexts or structured diagrams. GAOKAO-MM \cite{zong2024gaokao} and CMMMU \cite{ge2024cmmmu} reflect the format of high-stakes standardized tests, containing visually dense problem structures with minimal textual scaffolding. EMMA \cite{hao2025can} further enhances the realism of scientific instruction by combining textual questions with experimental setups or process diagrams, requiring temporal reasoning and procedural interpretation. To address the issue of visual leakage in multimodal evaluation datasets, MM-Star \cite{chen2024we} assesses the multimodal capabilities of LVLM by constructing a carefully balanced and purified sample selection process, with the aim of benchmarking six core competencies and 18 detailed axes.

\paragraph{Code Generation.}
Benchmarks in this category evaluate the ability of MLLMs to generate code based on multimodal visual prompts, such as plots or interface sketches. ChartMimic \cite{shi2024chartmimic} and Plot2Code \cite{wu2024plot2code} require models to produce plotting scripts such as matplotlib that replicate the semantics and structure of a given graph. These tasks test visual layout parsing, semantic inference, and parameter mapping between modalities. HumanEval-V \cite{zhang2024humaneval} extends this framework by offering open-ended coding tasks guided by visual problem descriptions, where models must synthesize intent, logic, and implementation through image-grounded reasoning.

\paragraph{Algorithmic Problem.}
The NPHardEval4V \cite{fan2024nphardeval4v} benchmark focuses on visually rendered algorithmic reasoning tasks, including classical problems such as graph coloring, knapsack optimization, and shortest path computation. The tasks require interpreting structured visual representations and aim to explore the model's ability to map visual configurations to algorithmic abstractions and perform reasoning within computationally complex solution spaces.

\paragraph{Social and Cultural Knowledge Reasoning.}
This subcategory evaluates MLLMs’ ability to reason about sociocultural, geographic, and symbolic content presented in visual form. II-Bench \cite{liu2024ii}, CII-Bench \cite{zhang2024can}, PunchBench \cite{ouyang2024punchbench}, and GeoComp \cite{song2025geolocation} introduce tasks involving photographs, illustrations, posters, and maps, where models must interpret not only the visual content but also its embedded cultural or societal implications. CII-Bench \cite{zhang2024can} features Chinese cultural imagery including traditional art and contemporary symbolism, probing metaphor understanding and emotional alignment. PunchBench \cite{ouyang2024punchbench} and GeoComp \cite{song2025geolocation} emphasize spatial-social reasoning grounded in world knowledge and demographic cues. These datasets evaluate whether models can go beyond surface-level perception to perform culturally and contextually sensitive reasoning.

\subsection{Contextual and Perceptual Reasoning in Real-World Scenarios}

This category focuses on benchmarks that simulate real-world scenarios where reasoning is grounded in contextual understanding and perceptual input. Tasks often involve dynamic visual contexts, user interfaces, or spatiotemporal environments that require MLLMs to interpret situational cues, predict goal-directed actions, or reason about spatial relationships. These benchmarks evaluate whether models can align visual perception, linguistic instructions, and environmental constraints to perform coherent reasoning in interactive or temporally evolving settings.

\paragraph{Action Prediction.} Benchmarks in this subcategory simulate scenarios where agents must make predictions or decisions based on dynamic visual contexts. Driving-Hazard-Prediction-and-Reasoning \cite{charoenpitaks2024exploring} presents real-world traffic images and sequential scenarios to assess hazard identification and behavioral prediction. InfiGUIAgent-Data \cite{liu2025infiguiagent}, Mind2Web \cite{deng2023mind2web}, and SeeClick \cite{cheng2024seeclick} extend this to user-interface environments, requiring models to reason about user intent and anticipate interactions with graphical user interfaces. 
Mind2Web \cite{deng2023mind2web} further introduces a broad set of high-level web-based tasks grounded in real-world websites, evaluating models on instruction following, multi-step planning, and interface understanding. 

\paragraph{Spatial Reasoning.} 
Spatial reasoning benchmarks target the model’s understanding of object layout, spatial relationships, and three-dimensional scene structure. iVISPAR \cite{mayer2025ivispar} evaluates MLLMs on spatial visual question answering tasks involving object localization, spatial relation inference, and layout understanding in cluttered indoor scenes. It emphasizes relational reasoning and occlusion modeling based on realistic 2D imagery. 
PulseCheck457 \cite{wang2025pulsecheck457} advances this work by introducing the first benchmark explicitly designed for 6D spatial reasoning. It assesses models on multi-object recognition, 2D and 3D positioning, and 3D orientation within a synthetic, unbiased environment, covering five difficulty levels and seven question types. This provides structured diagnostics for spatial abstraction capabilities in embodied robotics and AR/VR applications.
GeoSense \cite{xu2025geosense} further emphasizes symbolic-visual integration by evaluating geometric principle identification and application in diagram-based reasoning. VisuLogic \cite{xu2025visulogic} targets core visual reasoning skills, focusing on symmetry, rotation, folding, and positional changes in structured visual puzzles.

\subsection{Generalization and Multi-Task Reasoning Benchmarks}

This category includes composite benchmarks designed to assess MLLMs’ abilities in multi-task reasoning, cross-domain generalization, and zero-shot adaptation. These datasets typically aggregate diverse tasks across modalities, testing whether models can perform structured inference with minimal supervision. Tasks range from multimodal CoT generation to inductive abstraction and analogical reasoning, providing a comprehensive evaluation of models’ capacity to generalize beyond domain-specific constraints and perform flexible reasoning over heterogeneous inputs.

\paragraph{Multi-Image Based Inductive Reasoning.} To evaluate inductive abstraction and rule discovery, benchmarks like MM-IQ \cite{cai2025mm}, LogicVista \cite{xiao2024logicvista}, and PuzzleTest \cite{toh2025jumping} present visual matrix-style puzzles inspired by Raven’s Progressive Matrices. These benchmark challenge models to infer abstract transformations, detect visual analogies, and fill in missing patterns by synthesizing structural invariants across multiple image frames. They emphasize high-level visual abstraction and the ability to generalize across non-linguistic reasoning patterns.

\paragraph{Multi-Task and Generalization-Oriented Benchmarks.} 
A growing number of composite benchmarks have been proposed to evaluate generalized multimodal reasoning across varied tasks and domains. Representative datasets such as VRC-Bench \cite{thawakar2025llamav} and MLLM-COMPBench \cite{kil2024mllm} aggregate subtasks including visual QA, multi-hop reasoning, and tool-based inference, offering a unified setup for multi-skill evaluation. VCR-Bench \cite{qi2025vcr} further extends this line into the video domain by incorporating annotated CoT steps and assessing both answer and reasoning quality across temporal tasks. Notably, M3CoT \cite{chen2024m} and MME-CoT \cite{jiang2025mme} include multimodal CoT annotations, enabling fine-grained process supervision during inference. Meanwhile, ZeroBench \cite{roberts2502zerobench} targets zero-shot generalization across unseen tasks and modalities, testing model transferability and robustness without instruction tuning.
Together, these benchmarks offer a comprehensive suite for evaluating holistic, structured, and cross-modal reasoning under diverse and complex inputs.


\section{Limitations, Challenges and Future Directions}
\label{sec:limAndFuture}
\subsection{Limitations and Challenges}

\subsubsection{Limitations.} Despite the remarkable progress of RL-based reasoning methods in enhancing MLLMs, current research still faces multiple structural and theoretical limitations that hinder both generalization and scalability. 

\paragraph{Sparsity of reward signals.} One of the most pressing issues is the sparsity and delayed nature of reward signals. To avoid reward hacking, most RL-enhanced MLLMs are trained using scalar final-task rewards—such as answer correctness or classification accuracy—that only reflect task outcomes but offer no intermediate feedback throughout the reasoning process. This leaves the model blind to structural errors in earlier reasoning steps. The most immediate consequence of this issue is overthinking, where the model generates excessively long or redundant reasoning paths, with intermediate steps potentially including irrelevant visual or textual cues. Although recent works \cite{zhang2025r1, deng2025boosting} have introduced process-oriented feedback through intermediate step verification or curriculum-style stage partitioning, these designs are often handcrafted, task-specific, and difficult to generalize across unseen modalities or new domains.

\paragraph{Evaluation paradigms.} Another limitation lies in the benchmark-centric and static nature of current evaluation paradigms. Many RL-based MLLMs are trained and tested on a limited range of benchmarks, which are typically narrow in scope and fail to reflect the complexity and unpredictability of real-world multimodal reasoning. As a result, models trained on such benchmarks, especially smaller-scale ones, tend to exhibit poor transferability when applied to dynamic environments or introduced to new modalities like audio or 3D spatial layouts.

\paragraph{Real-time adaptivity and interactivity.} Furthermore, current approaches lack real-time adaptivity and interactivity. Most reinforcement signals are generated offline and assume static input-output mappings. In contrast, realistic deployments—such as embodied agents, web assistants, or interactive tutoring systems—require continuous feedback loops, the ability to revise reasoning, and responsiveness to user corrections. Without such interactive adaptability, many current MLLMs cannot bridge the gap between simulation-based training and open-world reasoning tasks.

\subsubsection{Challenges.} Beyond structural limitations, MLLM-RL pipelines face key challenges in optimization, multimodal alignment, and training infrastructure. A central issue is aligning diverse modalities such as images, text, audio, and spatial layouts under weak supervision. Real-world tasks often require complex cross-modal mappings with limited direct supervision, and designing reward functions to support consistent alignment, especially in open-ended scenarios, remains unsolved. Another challenge is modeling non-Markovian dependencies in reasoning trajectories. Unlike traditional RL, MLLM reasoning requires long-term consistency across modalities and steps. Although some work \cite{kang2025gflowvlm, pan2025metaspatial} has addressed this, optimization in non-Markovian spaces is prone to instability, noisy gradients, and unclear credit assignment. Finally, training-inference mismatch persists: models trained with fixed prompts and deterministic rewards are evaluated on unpredictable inputs, variable reasoning lengths, and ambiguous outputs—compromising real-world performance despite strong benchmarks.

\subsection{Future Directions}
To overcome these limitations and challenges, we propose several concrete directions for future research that can contribute to more robust, generalizable, and interactive RL in MLLMs:

\paragraph{Unified and hierarchical reward frameworks.} 
Future work should focus on designing multi-level reward signals that capture not only task outcomes but also the fidelity of reasoning and cross-modal coherence. For example, hybrid rewards can integrate accuracy, structure, and quality, thereby improving sample efficiency, interpretability, and policy stability.

\paragraph{Reward generalization across modalities.} To move beyond handcrafted or task-specific rewards, future models should adopt modular or learnable reward functions that generalize across images, videos, audio, and 3D modalities. Meta-learned reward functions or reward transformers that map trajectories to score distributions could reduce the reliance on extensive human supervision.
  
\paragraph{Scalable and lightweight reinforcement optimization.} A promising research direction is the development of lightweight RL techniques suitable for resource-constrained models. This includes curriculum learning, KL-regularized off-policy methods, and contrastive reward estimation. Such strategies can facilitate broader adoption in scenarios where deploying large models is infeasible.
  
\paragraph{Interactive reinforcement via user feedback.} Incorporating real-time user preferences, corrections, and demonstrations during test-time can help create continuously adapting MLLMs. Instead of relying on synthetic reward approximators, future systems could be equipped with interfaces for interactive reward elicitation, expanding the scope of RL in dynamic, user-centered settings.
  
\paragraph{Embodied and grounded multimodal environments.} Deploying MLLMs in spatially grounded environments such as robotics and AR/VR requires reinforcement learning strategies capable of handling physical constraints, causal relationships, and temporal dynamics. This includes integrating spatial consistency checks, collision detection, and user affordance modeling into the reward computation pipeline, as seen in MetaSpatial.

\section{Conclusion}

In this survey, we presented a comprehensive overview of RL-based reasoning methods in MLLMs. We analyzed recent progress in RL optimization strategies, categorized value-model-free and value-model-based approaches, and discussed reward mechanism designs that enable MLLMs to reason across text, vision, audio, and video modalities. By summarizing benchmark datasets, evaluation protocols, and practical applications, we provided a structured understanding of how RL is shaping the development of multimodal reasoning. While significant advancements have been made, RL-based multimodal reasoning still faces critical challenges. Current methods often rely on text-centric reasoning paths, underutilize multimodal information, and depend heavily on simple, verifiable reward designs. Furthermore, issues such as sparse and delayed rewards, inefficient exploration strategies, and limited generalization to open-ended tasks remain key obstacles. Looking ahead, future research should explore hierarchical and structured reward models, dynamic cross-modal CoT generation, and lightweight, scalable RL frameworks. Addressing these challenges is crucial for enabling MLLMs to perform more robust, interpretable, and generalized reasoning across real-world multimodal tasks. We hope this survey will serve as a valuable reference for researchers and practitioners, and inspire further innovation in this rapidly evolving field.


\bibliographystyle{ACM-Reference-Format}
\bibliography{sample-base}

\appendix


\end{document}